\documentclass[10pt,twocolumn,letterpaper]{article}

\usepackage{./ICCV2023/iccv}
\usepackage{times}
\usepackage{epsfig}
\usepackage{graphicx}
\usepackage{amsmath}
\usepackage{amssymb}

\usepackage{graphicx}
\usepackage{amsmath}
\usepackage{amssymb}
\usepackage{booktabs}

\usepackage{threeparttable}
\usepackage[vlined, ruled, linesnumbered]{algorithm2e}
\usepackage{multirow}
\usepackage{booktabs}
\usepackage{bbm}
\usepackage{pifont}
\usepackage{enumitem}

\usepackage{eucal,nicefrac}
\usepackage{subcaption}
\usepackage{gensymb}

\usepackage[margin=4pt,font=footnotesize,labelfont=bf,labelsep=endash,tableposition=top]{caption}

\usepackage[pagebackref=true,breaklinks=true,letterpaper=true,colorlinks,bookmarks=false]{hyperref}

\usepackage[capitalize]{cleveref}
\crefname{section}{Sec.}{Secs.}
\Crefname{section}{Section}{Sections}
\Crefname{table}{Table}{Tables}
\crefname{table}{Table}{Tables}

\iccvfinalcopy %

\def\iccvPaperID{5522} %

\ificcvfinal\fi

\begin{document}

\title{Metric3D: Towards Zero-shot Metric 3D Prediction from A Single Image}

\def\SP{~~}

\author{
Wei Yin$^{1 \ast}$,
\SP
Chi Zhang$^2$\thanks{Equal contributions.}, 
\SP 
Hao Chen$^3$\thanks{Corresponding author.},
\SP
Zhipeng Cai$^4$,
\SP 
Gang Yu$^2$,
\SP 
Kaixuan Wang$^1$, \\
\SP 
Xiaozhi Chen$^1$,
\SP 
Chunhua Shen$^3$
\\[0.1325cm]
$ ^1$ DJI Technology
\SP ~~~
$ ^2 $ Tencent
\SP ~~~
$ ^3$ Zhejiang University
\SP ~~~
$ ^4$ Intel Labs
\\
{e-mail: $\tt\small ^1 \{yvan.yin, halfbullet.wang, xiaozhi.chen\}@dji.com;$ }\\
$ \tt\small^2\{johnczhang, skicyyu\}@tencent.com;$ \\
$ \tt\small ^3 haochen.cad@zju.edu.cn, chunhua@me.com;$ 
$\tt\small ^4 zhipeng.cai@intel.com$ 
}

\makeatletter
\let\@oldmaketitle\@maketitle%
\renewcommand{\@maketitle}{\@oldmaketitle%
 \centering
    \includegraphics[width=0.95\textwidth]{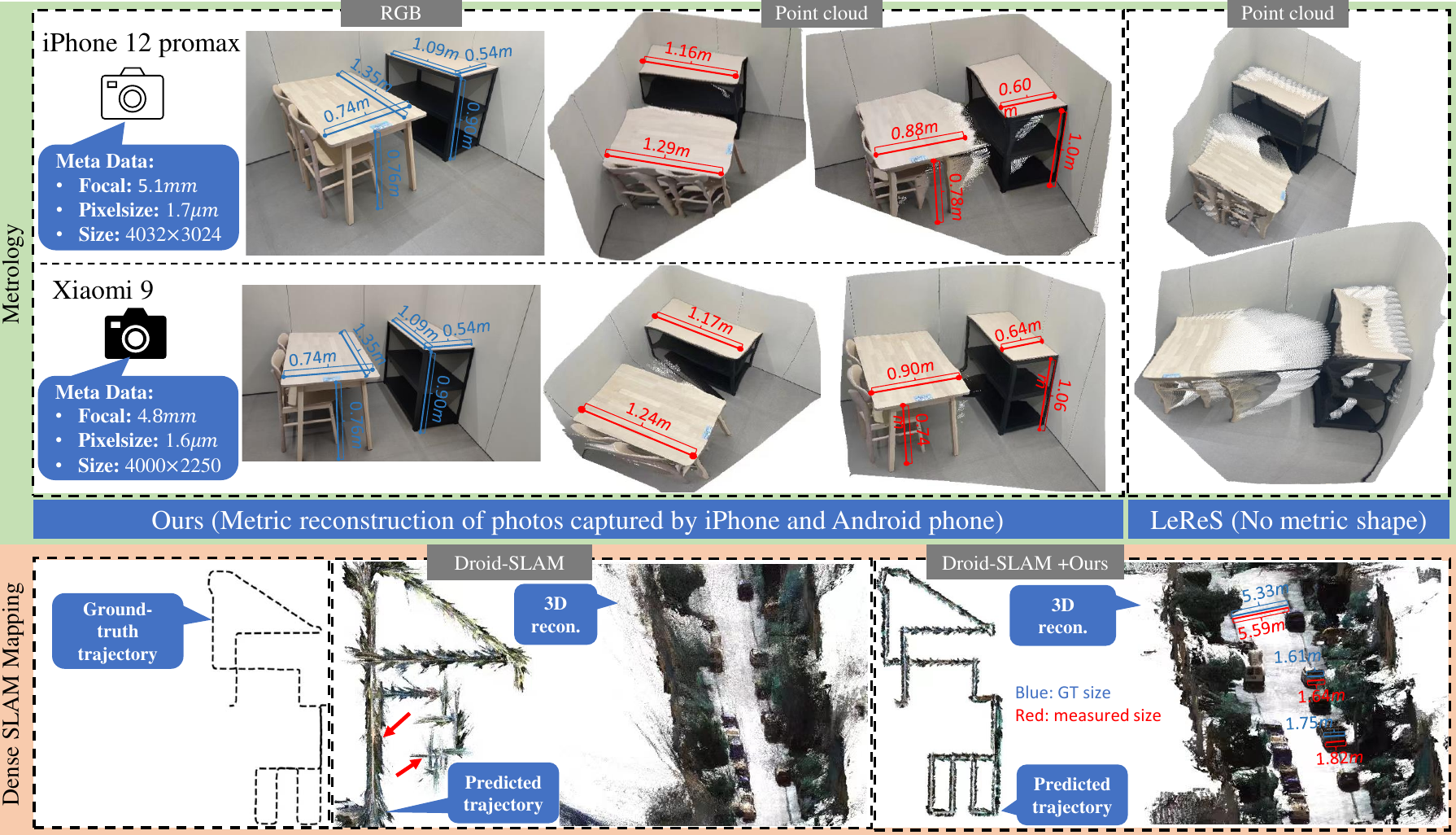}
     \captionof{figure}{\textbf{
     Illustration 
     and applications of our metric 3D reconstruction
     method}.
     Top (metrology):  we use two phones (iPhone 12 and an Android phone) to capture the scene and measure the size of tables. With the photos' metadata, we perform 3D metric reconstruction and then measure tables' sizes (marked in red), which are very close to the ground truth (marked in blue). In contrast, the recent method LeReS~\cite{leres} performs much worse and is unable to predict metric 3D by design.
     Bottom  (dense SLAM mapping): existing SOTA mono-SLAM methods usually face scale drift problems (see the red arrows) in large-scale scenes and are unable to achieve the metric scale, while, naively inputting our metric depth, Droid-SLAM~\cite{teed2021droid} can recover much more accurate trajectory and perform the \textit{metric} dense mapping (see the red measurements). 
     Note that all testing data are unseen to our model.
}
    \label{Fig: first page fig.}
    \bigskip}                   %
\makeatother

\maketitle

\def\PWN{{\rm PWN}}
\def\VNL{{\rm VNL}}
\def\RPNL{{\rm RPNL}}

\begin{abstract}

Reconstructing accurate 3D scenes from images is a long-standing vision task. Due to the ill-posedness of the single-image reconstruction problem, most well-established methods are built upon multi-view geometry. State-of-the-art (SOTA) monocular metric depth estimation methods can only handle a single camera model and are unable to perform mixed-data training due to the metric ambiguity. Meanwhile, SOTA monocular methods trained on large mixed datasets achieve zero-shot generalization by learning affine-invariant depths, which cannot recover real-world metrics. In this work, we show that the key to a zero-shot single-view metric depth model lies in the combination of large-scale data training and resolving the metric ambiguity from various camera models. We propose a canonical camera space transformation module, which explicitly addresses the ambiguity problems and can be effortlessly plugged into existing monocular models. Equipped with our module, monocular models can be stably trained over $8$ million of images with thousands of camera models, resulting in zero-shot generalization to in-the-wild images with unseen camera settings. 

\textbf{ Experiments demonstrate SOTA performance of our method on $7$ zero-shot benchmarks.
Notably, our method won the championship in the }\href{https://jspenmar.github.io/MDEC/}{2nd Monocular Depth Estimation Challenge}.
Our method enables the accurate recovery of metric 3D structures on randomly collected internet images, paving the way for plausible single-image metrology. The potential benefits extend to downstream tasks, which can be significantly improved by simply plugging in our model. 
For example, 
our model relieves the scale drift issues of monocular-SLAM (Fig.~\ref{Fig: first page fig.}), leading to high-quality metric scale dense mapping.  The code is available at \url{https://github.com/YvanYin/Metric3D}.
\end{abstract}

\section{Introduction}

3D reconstruction from images is the core of many computer vision applications, such as autonomous driving and robotics. Main-stream methods leverage multi-view %
geometry~\cite{hartley2003multiple} to confidently recover 3D structures. However, these methods cannot be applied to a single image, making 3D reconstruction hard without a prior. 
State-of-the-art transferable methods, such as MiDaS~\cite{Ranftl2020}, LeReS~\cite{leres}, and HDN~\cite{zhang2022hierarchical}, learn such a prior from a large dataset, but they can only output \emph{affine-invariant} depths, i.e., which are accurate only up to an unknown offset and scale. Though monocular metric depth estimation methods~\cite{yuan2022new, bhat2021adabins} work on a single dataset with a single camera model, they cannot generalize to unseen cameras or scenes. 
This work aims to address the above problems by learning a \emph{zero-shot, single view, metric} depth model.

According to the predicted depth, 
existing methods are categorized into learning metric depth \cite{yuan2022new, yin2021virtual, bhat2021adabins, yang2021transformers}, learning relative depth \cite{xian2018monocular, xian2020structure, chen2020oasis, chen2016single}, and learning affine-invariant depth~\cite{leres, yin2022towards, Ranftl2020, ranftl2021vision, zhang2022hierarchical}. Although the metric depth methods~\cite{yuan2022new, yin2021virtual, Yin2019enforcing, bhat2021adabins, yang2021transformers} have achieved impressive accuracy on various benchmarks, they must train and test on the dataset with the same camera intrinsics. %
Therefore, the training datasets of metric depth methods are often small, as it is hard to collect a large dataset covering diverse scenes using one identical camera. The consequence is that 
all these models are not transferable -- they generalize poorly to images in the wild, not to mention the camera parameters of test images can vary too.
A compromise is to learn the relative depth
\cite{chen2020oasis, xian2018monocular}, which only represents
one point being further or closer to another one. %
The application of relative depth is very limited.
Learning affine-invariant depth 
finds a trade-off between the above two categories of methods. 
With large-scale data, they decouple the metric information during training and achieve impressive robustness and generalization ability. The recent state-of-the-art LeReS~\cite{leres} can recover 3D scenes in the wild, 
but only up to an unknown scale and shift. 

This work focuses on learning a zero-shot transferable model to recover \textit{metric} 3D from a single image.
First, we  analyze the metric ambiguity issues in monocular depth estimation and study different camera parameters in depth, including the pixel size, focal length, and sensor size. We observe that 
the focal length is the critical factor for accurate metric recovery. %
By design, LeReS~\cite{leres} 
does not take the focal length information into account during training. As shown in Sec.~\ref{sec:ambiguity}, only from the image appearance, various focal lengths may cause metric ambiguity, thus they decouple the depth scale in training.
To solve the problem of varying focal lengths, CamConv~\cite{facil2019cam}  encodes the camera model in the network, which enforces the network to implicitly understand camera models from
the image appearance and then bridges the imaging size to
the real-world size. However, training data contains limited images and types of cameras, %
which challenges data diversity and network capacity.
In contrast, we propose a canonical camera transformation method in training. It is inspired by the human body reconstruction methods. To improve reconstructed shape quality on countless poses, they map all samples to a canonical pose space~\cite{peng2022animatable} to reduce pose variance. Similarly, 
we transform all training data to a canonical camera space where the processed images are coarsely regarded as captured by the same camera.
To achieve such transformation, we propose two different methods. The first one tries to adjust the image appearance to simulate the canonical camera, while the other one transforms the ground-truth labels for supervision. Camera models are not encoded in the network, making our method easily applicable to existing architectures. During inference, a de-canonical transformation is employed to recover metric information. 
To further boost the depth accuracy, we propose a random proposal normalization loss. It is inspired by the scale-shift invariant loss~\cite{leres, Ranftl2020,zhang2022hierarchical}, which decouples the depth scale to emphasize the single image's distribution. However, they perform on the whole image, which inevitably squeezes the fine-grained depth difference. We propose to randomly crop several patches from images and enforce the scale-shift invariant loss~\cite{leres, Ranftl2020} on them. Our loss emphasizes the local geometry and distribution of the single image. 

With the proposed method, we can easily scale up model training to \emph{8 million} images from 11 datasets of diverse scene types (indoor and outdoor) and camera models (tens of thousands of different cameras), leading to zero-shot transferability and a significantly improved accuracy. %
Our model can accurately reconstruct metric 3D from randomly collected Internet images, enabling plausible single-image metrology. 
Different from affine-invariant depth models, our model can also directly improve various downstream tasks.
As an example (Fig.~\ref{Fig: first page fig.}), with the predicted metric depths from our model, we can significantly reduce the scale drift of monocular SLAM~\cite{teed2021droid, sun_2022_TRO} systems, achieving much better mapping quality with \emph{real-world metric recovery}. Our model also enables large-scale 3D reconstruction~\cite{im2019dpsnet}. \textbf{ The model achieves the championship in the 2nd Monocular Depth Estimation Challenge}~\cite{Spencer_2023_CVPR}.
To summarize, our main contributions are:
\begin{itemize}
\itemsep -0.1cm 
    \item We propose a canonical and de-canonical camera transformation method to solve the metric depth ambiguity problems from various cameras setting. It enables the learning of strong zero-shot monocular metric depth models from large-scale datasets. %
    \item We propose a random proposal normalization loss to effectively boost the depth accuracy;
    \item Our model achieves state-of-the-art performance on $7$ zero-shot benchmarks. It can perform high-quality 3D metric structure recovery in the wild and benefit several downstream tasks, such as mono-SLAM~\cite{teed2021droid, mur2017orb}, 3D scene reconstruction~\cite{im2019dpsnet}, and metrology~\cite{zhu2020single}.  
\end{itemize}

\section{Related Work}
\noindent\textbf{3D reconstruction from a single image.} 
Reconstructing various objects from a single image has been well studied~\cite{barron2014shape, wang2018pixel2mesh, wu2018learning}. They can produce high-quality 3D models of cars, planes, tables, and human body~\cite{saito2019pifu, saito2020pifuhd}. The main challenge is how to best recover objects' details, how to represent them with limited memory, and how to generalize to more diverse objects. However, all these methods rely on learning priors specific to a certain object class or instance, typically from 3D supervision, and can therefore not work for full scene reconstruction. Apart from these reconstructing objects works, several works focus on scene reconstruction~\cite{Xu_2023_ICCV} from a single image. Saxena~\etal~\cite{saxena2008make3d} construct the scene based on the assumption that the whole scene can be segmented into several small planes. With planes' orientation and location, the 3D structure can be represented. Recently, LeReS~\cite{leres} propose to use a strong monocular depth estimation model to do scene reconstruction. However, they can only recover the shape up to a scale. 
Zhang~\etal~\cite{Zhang_2023_ICCV} recently  propose a zero-shot geometry-preserving depth estimation model that is capable of making depth predictions up to an unknown scale, without requiring  scale-invariant depth annotations for training. 
In contrast to these works, our method can recover the metric 3D structure.

\noindent\textbf{Supervised monocular depth estimation.}
After several benchmarks~\cite{silberman2012indoor, Geiger2013IJRR} are established, neural network based
methods~\cite{yuan2022new, Yin2019enforcing, bhat2021adabins} have dominated since then. Several approaches regress the continuous depth from the aggregation of information in an image~\cite{eigen2014depth}. As depth distribution corresponding to different RGBs can vary to a large extent, some methods~\cite{Yin2019enforcing, 
bhat2021adabins}
discretize the depth and formulate this problem to a classification~\cite{yin2021virtual},  
which often achieves better performance. %
The generalization issue of deep models for 3D metric recovery   
is related to two problems. 
The first one is to generalize to diverse scenes, while 
the other 
one is how to predict accurate metric information under various camera settings. The first problem has been well addressed by recent methods. Some works~\cite{
xian2020structure, xian2018monocular, yin2021virtual} propose to construct a large-scale relative depth dataset, such as DIW~\cite{chen2016single} and OASIS~\cite{chen2020oasis}, and then they target learning the relative relations. However, the relative depth loses geometric structure information.
To improve the recovered geometry quality, 
learning affine-invariant depth methods, such as MiDaS~\cite{Ranftl2020}, LeReS~\cite{leres}, and HDN~\cite{ 
zhang2022hierarchical} are proposed. 
By mixing large-scale data, 
state-of-the-art performance and the generalization over scenes are improved continuously. 
Note that by design, these methods are unable to recover 
the metric information.
How to achieve both strong generalization and accurate metric information over diverse scenes is the key problem that
we attempt to tackle. 

\noindent\textbf{Large-scale data training.}
Recently, various natural language problems and computer vision problems~\cite{
yin2022devil, radford2021learning, lambert2020mseg} have achieved impressive progress with large-scale data training. CLIP~\cite{radford2021learning} is a promising classification model, which is trained on billions of paired image and language descriptions data. It achieved state-of-the-art performance over several classification benchmarks by zero-shot testing. 
For depth prediction, large-scale data training has been widely applied. Ranft~\etal~\cite{Ranftl2020} mixed over 2 million data in training, LeReS~\cite{yin2022towards} collected over $300$ thousands data, Eftekhar~\etal~\cite{eftekhar2021omnidata} also merged millions of data to build a strong depth prediction model. %

\section{Method}

\begin{figure*}[]
\centering
\includegraphics[width=0.99\textwidth]{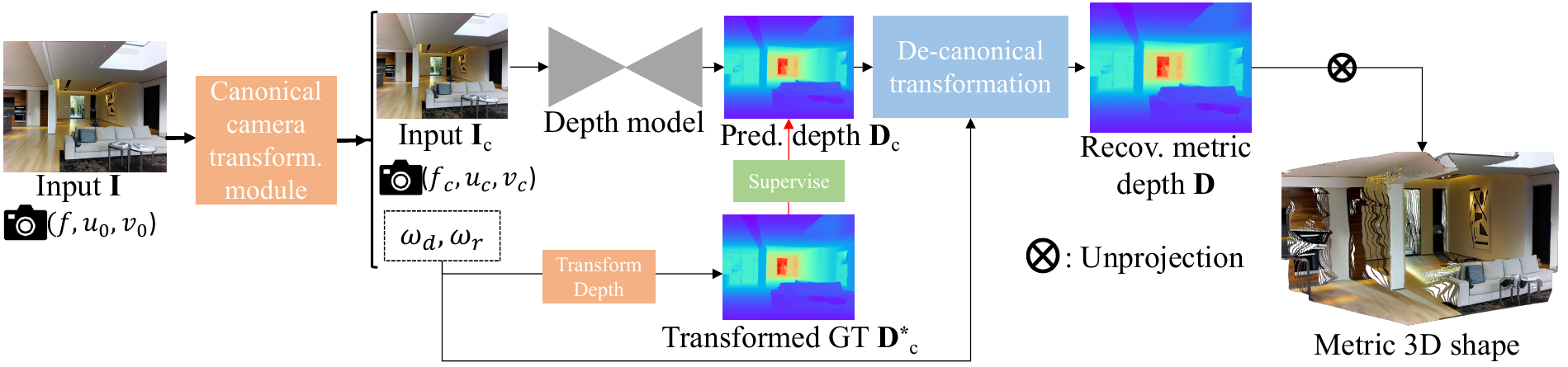}
\vspace{-1 em}
\caption{\textbf{Pipeline.} 
Given an input image $I$, we first transform it to the canonical space using CSTM. The transformed image $I_c$ is fed into a standard depth estimation model to produce the predicted metric depth $D_c$ in the canonical space. During training, $D_c$ is supervised by a GT depth $D^*_c$ which is also transformed into the canonical space. In inference, after producing the metric depth $D_c$ in the canonical space, we perform a de-canonical transformation to convert it back to the space of the original input $I$. The canonical space transformation and de-canonical transformation are executed using camera intrinsics.}
\label{fig: pipeline}
\vspace{-1em}
\end{figure*}

\noindent\textbf{Preliminaries.}
We consider the pin-hole camera model with intrinsic parameters are:
[[$\nicefrac{\hat{f}}{\delta}, 0, u_{0}$], [$0,  \nicefrac{\hat{f}}{\delta},  v_{0}$], [$0, 0,  1$]],
where $\hat{f}$ is the 
focal length (in micrometers), $\delta$ is the pixel size
(in micrometers), and
$(u_{0}, v_{0})$ is the principle center. $f = \nicefrac{\hat{f}}{\delta}$ is the pixel-represented focal length used in vision algorithms.

\subsection{Metric Ambiguity Analysis}\label{sec:ambiguity}

\begin{figure}[!bt]
\centering
\includegraphics[width=0.47\textwidth]{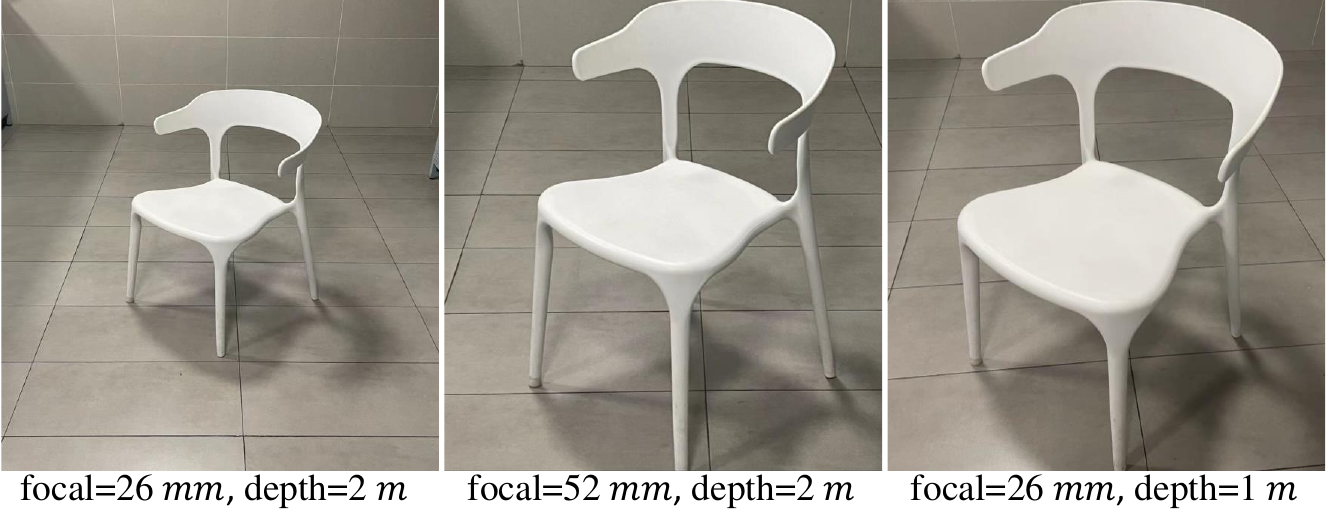}
\vspace{-0.5 em}
\caption{\textbf{Photos of a chair captured at different distances with different cameras}. The first two photos are captured at the same distance but with different cameras, 
while the last one is taken at a closer distance with the same camera as the first one.}
\label{fig: inspiration}
\vspace{-1.5em}
\end{figure}

Fig.~\ref{fig: inspiration} presents an example of photos taken by different cameras and at 
different distances. Only from the image's appearance, 
one may think 
the last two photos are taken at %
a 
similar location by the same camera. %
In fact, due to different focal lengths, 
these %
are captured at different locations.
Thus, 
camera %
intrinsic parameters are 
critically 
important for the metric estimation from a single image, as otherwise, the problem is \textit{ill posed}. 
To avoid such metric ambiguity, recent %
methods, such as MiDaS~\cite{Ranftl2020} and LeReS~\cite{leres}, decouple the metric from the supervision and %
compromise learning the affine-invariant depth.

\begin{figure}[!b]
\vspace{-2em}
\centering
\includegraphics[width=0.5\textwidth]{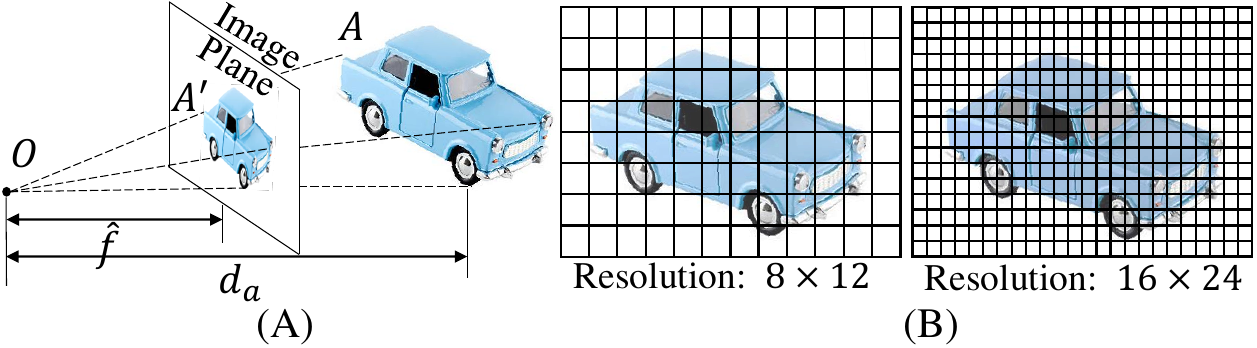}
\caption{\textbf{Pinhole camera model}. (A) Object $A$ at the distance $d_{a}$ is projected to the image plane. (B) Using two cameras to capture the car. The left one has a larger pixel size. Although the projected imaging sizes are the same, the pixel-represented images (resolution) are different.}
\label{fig: pinhole camera}
\vspace{-0.5em}
\end{figure}

Fig.~\ref{fig: pinhole camera} (A) shows a simple pin-hole perspective projection. Object $A$ locating at $d_{a}$ is projected to $A'$. 
Based on the principle of similarity, we %
have the equation:
\begin{equation}
\vspace{-1em}
    d_{a} = \hat{S} \Bigl[\frac{\hat{f}}{\hat{S'}}\Bigr]= \hat{S}\cdot \alpha
\label{eq: similarity}
\end{equation}
where $\hat{S}$ and $\hat{S'}$ are the real and \textit{imaging} size respectively. $\hat{\cdot}$ denotes variables are in the physical metric (\textit{e.g.}, millimeter). To %
recover 
$d_{a}$ from a single image, focal length, imaging size of the object, and real-world object size %
must be available. 
Estimating the focal length 
from a single image is a %
challenging 
and ill-posed problem. Although several methods~\cite{leres, hold2018perceptual} have explored,
the accuracy 
is still far from being satisfactory. 
Here, we simplify the problem by assuming 
the focal length of a training/test image is available. 
In contrast, understanding the imaging size is much easier for a neural network. To obtain the real-world object size, a neural network %
needs to 
understand the semantic scene layout and the object, at which a neural network excels. 
We %
define 
$\alpha = \nicefrac{\hat{f}}{\hat{S'}} $, so $d_{a}$ is proportional to $\alpha$. 

We make the following observations regarding sensor size, pixel size, and focal length.

\noindent\textbf{O1: Sensor size and pixel size do not affect the metric depth estimation.} 
Based on the perspective projection (Fig.~\ref{fig: pinhole camera} (A)), the sensor size only affects the field of view (FOV) and is irrelevant to $\alpha$, thus does not affect the metric depth estimation. 
For the pixel size, %
we assume two cameras with different pixel sizes ($\delta_{1} = 2\delta_{2}$) but the same focal length $\hat{f}$ to capture the same object locating at $d_{a}$. %
Fig.~\ref{fig: pinhole camera} (B) shows their captured photos.
According to the preliminaries, %
the pixel-represented focal length $f_{1} = \frac{1}{2} f_{2}$. 
As the second camera has a smaller pixel size, although in the same projected imaging size $\hat{S'}$, the pixel-represented image resolution is $S'_{1} = \frac{1}{2} S'_{2}$. According to Eq.~\eqref{eq: similarity}, $\frac{\hat{f}}{\delta_{1}\cdot S'_{1}} = \frac{\hat{f}}{\delta_{2}\cdot S'_{2} }$, i.e. $\alpha_1 = \alpha_2$, so $d_{1} = d_{2}$. Therefore, different camera sensors %
would 
not affect the metric depth estimation.

\begin{figure}[!bt]
\centering
\includegraphics[width=0.48\textwidth]{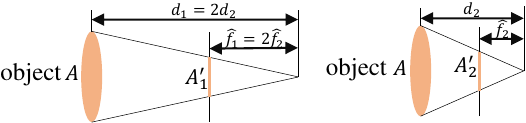}
\caption{\textbf{Illustration of two cameras with different focal length} at different distance. As $f_1=2f_2$ and $d_1=2d_2$, 
$A$ is projected 
to two image planes with the same imaging size (i.e. $A^{'}_1 = A^{'}_2$).
}
\label{fig: confusion}
\vspace{-2em}
\end{figure}

\noindent\textbf{O2: The focal length is vital for %
metric depth estimation}. Fig.~\ref{fig: inspiration} shows the metric ambiguity issue caused by the unknown focal length. 
Fig.~\ref{fig: confusion} %
illustrates 
this. If two cameras ($\hat{f}_{1} = 2\hat{f}_{2}$) are at distances $d_{1} = 2d_{2}$, the imaging sizes on cameras are the same. Thus, only from the appearance, %
the network will be confused when supervised with different labels.
Based on this observation, we propose a canonical camera transformation method to solve the supervision and image appearance conflicts.

\subsection{Canonical Camera Transformation}
The core idea is to set up a canonical camera space ($(f_{x}^{c}, f_{y}^{c})$, $f_{x}^{c}=f_{y}^{c}=f^{c}$ in experiments) and transform all training data to this space. 
Consequently, all data can roughly be regarded as captured by the canonical camera.
We propose two transformation methods, i.e. either transforming the input image ($\mathbf{I}\in\mathbb{R}^{H \times W \times 3}$) or the ground-truth (GT) label ($\mathbf{D}\in\mathbb{R}^{H \times W}$). The original intrinsics are $\{f, u_{0}, v_{0}\}$.

\noindent\textbf{Method1: transforming depth labels (CSTM\_label).}
Fig.~\ref{fig: inspiration}'s ambiguity is for depths. 
Thus our first method directly transforms the ground-truth depth labels to solve this problem.  Specifically, we scale the ground-truth depth ($\mathbf{D}^{*}$) with the ratio $\omega_d = \frac{f^{c}}{f}$ in training, \textit{i.e.},  $\mathbf{D}^{*}_{c} = \omega_d \mathbf{D}^{*}$. The original camera model is transformed to $\{f^{c}, u_{0}, v_{0}\}$. In inference, the predicted depth ($\mathbf{D}_{c}$) is in the canonical space and needs to perform a de-canonical transformation to recover the metric information, \textit{i.e.}, $\mathbf{D} = \frac{1}{\omega_d}\mathbf{D}_{c}$. Note the input $\mathbf{I}$ does not perform any transformation, \textit{i.e.},  $\mathbf{I}_c = \mathbf{I}$.

\noindent\textbf{Method2: transforming input images (CSTM\_image).}
From another view, the ambiguity is caused by the similar image appearance. Thus this method is to transform the input image to simulate the canonical camera imaging effect.
Specifically, the image $\mathbf{I}$ is resized with the ratio $\omega_r=\frac{f^{c}}{f}$, \textit{i.e.},
$\mathbf{I}_{c} = \mathcal{T}(\mathbf{I}, \omega_r)$, where $\mathcal{T}(\cdot)$ denotes image resize. The optical center is resized, thus the canonical camera model is $\{f^{c}, \omega_r u_{0}, \omega_r v_{0}\}$. The ground-truth labels are resized without any scaling, \textit{i.e.}, 
$\mathbf{D}^{*}_{c} = \mathcal{T}(\mathbf{D}^*, \omega_r)$. In inference, the de-canonical transformation is to resize the prediction to the original size without scaling, \textit{i.e.}, $\mathbf{D} = \mathcal{T}(\mathbf{D}_{c}, \frac{1}{\omega_r})$.

Fig.~\ref{fig: pipeline} shows the pipeline. After performing either transformation, we randomly crop a patch for training. %
The cropping only adjusts the FOV and the optical center, %
thus not causing any metric ambiguity issues. In the labels transformation method $\omega_r = 1$ and $\omega_d=\frac{f^c}{f}$, while $\omega_d = 1$ and $\omega_r=\frac{f^c}{f}$ in the images transformation method. The training objective is as follows:
\begin{equation}
    \min_{\theta}\left | \mathcal{N}_{d}(\mathbf{I}_{c}, \theta) - \mathbf{D}^{*}_{c} \right| 
\label{eq: robust metric depth}
\vspace{-0.5 em}
\end{equation}
where 
$\theta$ is the network's ($\mathcal{N}_{d}(\cdot)$) parameters, $\mathbf{D}^{*}_{c}$ and $\mathbf{I}_{c}$ are transformed ground-truth depth labels and images.

Mix-data training is an effective way to boost generalization. 
We collect $11$ datasets for training, see 
the supplementary materials for details. In the mixed data, over 10K different cameras are included. %
All collected training data have included paired camera intrinsic parameters, which are %
used in our canonical transformation module.

\noindent\textbf{Supervision.} To further boost the performance, we propose a random proposal normalization loss (RPNL). The scale-shift invariant loss~\cite{Ranftl2020, leres} is widely applied for the affine-invariant depth estimation, which decouples the depth scale to emphasize the single image distribution. However, such normalization based on the whole image inevitably squeezes the fine-grained depth difference, particularly in close regions. Inspired by this, we propose to randomly crop several patches ($p_{i(i=0,...,M)} \in \mathbb{R}^{h_i \times w_i}$) from the ground truth $\mathbf{D}^{*}_c$ and the predicted depth $\mathbf{D}_c$. Then we employ the median absolute deviation normalization~\cite{singh2019investigating} for paired patches. By normalizing the local statistics, we can enhance local contrast. The loss function is as follows:
\begin{eqnarray}\nonumber
    L_{\RPNL} = \frac{1}{MN} \sum_{p_i}^{M}\sum_{j}^{N} \lvert \frac{d^{*}_{p_i, j} - \mu(d^{*}_{p_i, j})}{\frac{1}{N}\sum_{j}^{N} \left |d^{*}_{p_i, j} - \mu(d^{*}_{p_i, j}) \right |} - \\
    \frac{d_{p_i, j} - \mu(d_{p_i, j})}{\frac{1}{N}\sum_{j}^{N} \left | d_{p_i, j} - \mu(d_{p_i, j}) \right |} \rvert
\label{eq: RPNL}
\end{eqnarray}
where $d^*\in \mathbf{D}^*_c$ and $d \in \mathbf{D}_c$ are the ground truth and predicted depth respectively. $\mu(\cdot)$ and is the median of depth. $M$ is the number of proposal crops, which is set to 32. During training, proposals are randomly cropped from the image by 
$0.125$ to $0.5$ of the original size. Furthermore, several other losses are employed, including the scale-invariant logarithmic loss~\cite{eigen2014depth} $L_{silog}$, pair-wise normal regression loss~\cite{leres}$L_{\PWN}$, virtual normal loss~\cite{yin2021virtual} $L_{\VNL}$. Note $L_{silog}$ is a variant of L1 loss.  The overall losses are as follows.
\begin{eqnarray}\nonumber
    L = L_{\PWN} + L_{\VNL} + L_{silog} + L_{\RPNL}.
\label{eq: losses}
\end{eqnarray}
\vspace{-2 em}

\section{Experiments}

\noindent\textbf{Dataset details.}
\label{sec:data}
We collect $11$ public RGB-D datasets, and over $8$ million data for training. It spreads over diverse indoor and outdoor scenes. Note that all datasets have provided %
camera intrinsic parameters. Apart from the test split of training datasets, we collect $7$ unseen datasets for robustness and generalization evaluation. Details of employed data are reported in the supplementary materials.

\noindent\textbf{Implementation details.}
We employ an UNet architecture with the ConvNext-large~\cite{liu2022convnet} backbone. ImageNet-22K pre-trained weights are used for initialization. We use AdamW with a batch size of $192$, an initial learning rate $0.0001$ for all layers, and the polynomial decaying method with the power of $0.9$. We train our final model on $48$ A100 GPUs for $500$K iterations. Following the DiverseDepth~\cite{yin2021virtual}, we balance all datasets in a mini-batch to ensure each dataset accounts for an almost equal ratio. During training, images are processed by the canonical camera transformation module, flipped horizontally with a $50\%$ chance, and then randomly cropped into 
$512 \times 960$ pixels. For the ablation experiments, training settings are different as we sample $5000$ images from each dataset for training. We trained on $8$ GPUs for $150$K iterations.

\noindent\textbf{Evaluation details.}
a) To show the robustness of our metric depth estimation 
method, we test on 8 zero-shot benchmarks, including NYUv2~\cite{silberman2012indoor}, KITTI~\cite{Geiger2013IJRR}, NuScenes~\cite{caesar2020nuscenes}, 7-scenes~\cite{shotton2013scene}, iBIMS-1~\cite{koch2018evaluation}, DIODE~\cite{vasiljevic2019diode}, ETH3D~\cite{schops2017multi}. Following previous works~\cite{yuan2022new}, absolute relative error (AbsRel),  the accuracy under threshold ($\delta_{i} < 1.25^{i}, i=1, 2, 3$), root mean squared error (RMS), root mean squared error in log space (RMS\_{log}), and log10 error (log10) metrics are employed. 
b) Furthermore, %
we also follow current affine-invariant depth benchmarks~\cite{leres, zhang2022hierarchical} (Tab. \ref{Table: generalization evaluation.}) to evaluate the generalization ability on $5$ zero-shot datasets, \textit{i.e.},  NYUv2, DIODE, ETH3D, ScanNet~\cite{dai2017scannet}, and KITTI. We mainly compare with large-scale data trained models. Note that in this benchmark we follow existing methods to apply the scale shift alignment before evaluation. 
c) To evaluate our metric 3D reconstruction quality, we randomly sample 9 unseen scenes from NYUv2 and use colmap~\cite{schoenberger2016mvs} to obtain the camera poses for multi-frame reconstruction. Chamfer $l_1$ distance and the F-score~\cite{knapitsch2017tanks} are used to evaluate the reconstruction accuracy. 
d) In dense-SLAM experiments, following Li~\etal~\cite{li2021generalizing}, we test on the KITTI odometry benchmark~\cite{Geiger2013IJRR} and evaluate the average translational RMS drift ($\%, t_{rel}$) and rotational RMS drift ($\degree/100m, r_{rel}$) errors~\cite{Geiger2013IJRR}.
Note that all these depth and reconstruction evaluations use the same trained model. 

\subsection{Zero-shot Generalization %
Test
}

\begin{table}[!t]
\caption{Quantitative comparison on NYUv2 and KITTI benchmarks. Both datasets are unseen to our model, but we can achieve comparable performance with state-of-the-art methods.}
\vspace{-1 em}
\scalebox{0.67}{
\begin{tabular}{r |cccccc}
\toprule[1pt]
\multicolumn{7}{c}{\textbf{NYUv2 Benchmark}} \\ \hline
\multirow{1}{*}{Method} & $\boldsymbol{\delta_{1}}$$\uparrow$ & $\boldsymbol{\delta_{2}}$$\uparrow$ & $\boldsymbol{\delta_{3}}$$\uparrow$ & \textbf{AbsRel}$\downarrow$ & \textbf{log10}$\downarrow$ & \textbf{RMS}$\downarrow$  \\ \hline
Li \etal.~\cite{li2017two}               & $0.788$    & $0.958$    & $0.991$  & $0.143$   & $0.063$    & $0.635$     \\
Laina \etal.~\cite{laina2016deeper}      & $0.811$    & $0.953$    & $0.988$  & $0.127$   & $0.055$    & $0.573$       \\
VNL ~\cite{Yin2019enforcing}            & $0.875$   & $0.976$    & $0.994$  & $0.108$   & $0.048$    & $0.416$    \\ 
TrDepth~\cite{yang2021transformers}     & $0.900$   & $0.983$    & $0.996$  & $0.106$  & $0.045$     & $0.365$   \\
Adabins~\cite{bhat2021adabins}         & $0.903$    & ${0.984}$  & $\underline{0.997}$  & $0.103$  & $0.044$     & $0.364$    \\
NeWCRFs~\cite{yuan2022new}              
& ${0.922}$  & $\boldsymbol{0.992}$  & $\boldsymbol{0.998}$ 
& $0.095$  & $0.041$     & $\underline{0.334}$   \\ \hline
Ours CSTM\_image    
& $\boldsymbol{0.925}$  & $0.983$  & $0.994$ 
& $\underline{0.092}$   & $\underline{0.040}$   & ${0.341}$  \\
Ours CSTM\_label    
& $\underline{0.944}$  & $\underline{0.986}$  & $0.995$ 
& $\boldsymbol{0.083}$   & $\boldsymbol{0.035}$   & $\boldsymbol{0.310}$  \\\hline
\hline
\multicolumn{7}{c}{\textbf{KITTI Benchmark}} \\ \hline \hline
\multirow{1}{*}{Method} & $\boldsymbol{\delta_{1}}$$\uparrow$ & $\boldsymbol{\delta_{2}}$$\uparrow$ & $\boldsymbol{\delta_{3}}$$\uparrow$ & \textbf{AbsRel} $\downarrow$ & \textbf{RMS} $\downarrow$ & \textbf{RMS\_log} $\downarrow$ \\ \hline
Guo \etal \cite{guo2018learning}  & $0.902$  & $0.969$  & $0.986$ & $0.090$ & $3.258$ & $0.168$    \\
VNL~\cite{Yin2019enforcing} & ${0.938}$   & ${0.990}$   & ${0.998}$  & ${0.072}$  & $3.258$      & ${0.117}$    \\ 
TrDepth~\cite{yang2021transformers}   & $0.956$  & $0.994$  & $0.999$   & $0.064$  & $2.755$  & $0.098$  \\
Adabins~\cite{bhat2021adabins} & $0.964$  & $0.995$  & $0.999$   & $\underline{0.058}$  & $2.360$  & $0.088$  \\
NeWCRFs~\cite{yuan2022new} 
& $\boldsymbol{0.974}$  & $\boldsymbol{0.997}$  & $\underline{0.999}$   & $\boldsymbol{0.052}$  & $\boldsymbol{2.129}$  & $\boldsymbol{0.079}$  \\ \hline
Ours CSTM\_image & 
$\underline{0.967}$   & $\underline{0.995}$   & $\boldsymbol{0.999}$  
& $0.060$  & ${2.843}$     & $\underline{0.087}$    \\ 
Ours CSTM\_label 
& ${0.964}$   & ${0.993}$   & ${0.998}$  
& $0.058$  & ${2.770}$     & ${0.092}$    \\ \hline
\toprule[1pt]
\end{tabular}\newline}
\label{table:errors cmp on NYUD-V2}
\vspace{-2 em}
\end{table}

\begin{table*}[]
\renewcommand\arraystretch{1.1}
\caption{Quantitative comparison of 3D scene reconstruction with LeReS~\cite{leres}, DPT~\cite{ranftl2021vision}, %
RCVD~\cite{kopf2021rcvd}, 
SC-DepthV2~\cite{bian2021tpami}, and a learning-based MVS method (DPSNet~\cite{im2019dpsnet}) on 9 unseen NYUv2 scenes. Apart from DPSNet and ours, other methods have to align the scale with ground truth depth for each frame. As a result, our reconstructed 3D scenes achieve the best performance.}
\vspace{-1 em}
\centering
\resizebox{.98\linewidth}{!}{%
  \centering
  \small 
  \setlength{\tabcolsep}{0.5mm}{\begin{tabular}{@{} r |rc|rc|rc|rc|rc|rc|rc|rc|rc@{}}
    \toprule
    \multirow{2}{*}{Method} & \multicolumn{2}{c|}{Basement\_0001a} & \multicolumn{2}{c|}{Bedroom\_0015} & \multicolumn{2}{c|}{Dining\_room\_0004} & \multicolumn{2}{c|}{Kitchen\_0008} & \multicolumn{2}{c|}{Classroom\_0004} & \multicolumn{2}{c|}{Playroom\_0002}  & \multicolumn{2}{c|}{Office\_0024} & \multicolumn{2}{c|}{Office\_0004} & \multicolumn{2}{c}{Dining\_room\_0033}\\
      & C-$l_1$$\downarrow$ & F-score $\uparrow$ & C-$l_1$$\downarrow$ & F-score $\uparrow$ & C-$l_1$$\downarrow$ & F-score $\uparrow$ & C-$l_1$$\downarrow$ & F-score $\uparrow$ & C-$l_1$$\downarrow$ & F-score $\uparrow$ & C-$l_1$$\downarrow$ & F-score $\uparrow$ &
      C-$l_1$$\downarrow$ & F-score $\uparrow$ & C-$l_1$$\downarrow$ & F-score $\uparrow$ & C-$l_1$$\downarrow$ & F-score $\uparrow$\\ \hline
    RCVD~\cite{kopf2021rcvd} & 0.364 & 0.276 
            & 0.074 & 0.582 &
             0.462 & 0.251 &
             0.053 & 0.620 &
              0.187 & 0.327 &
             0.791 & 0.187 &
             0.324 & 0.241  &
             0.646 & 0.217 &
             0.445 & 0.253 \\

    SC-DepthV2~\cite{bian2021tpami}  & 0.254 & 0.275 &
             0.064 & 0.547 &
             0.749 & 0.229 &
             0.049 & 0.624 &
              0.167 & 0.267 &
             0.426 & 0.263 &
             0.482 & 0.138  &
             0.516 & 0.244 &
             0.356 &0.247 \\

    DPSNet~\cite{im2019dpsnet} & 0.243 & 0.299 &
             0.195 & 0.276 &
             0.995 & 0.186 &
             0.269 & 0.203 &
             0.296 & 0.195 &
             0.141 & 0.485 &
             0.199 & 0.362  &
             0.210 & 0.462 &
             0.222 & 0.493 \\
              
    DPT~\cite{leres} & 0.698 & 0.251 &
             0.289 & 0.226 &
             0.396 & 0.364 &
             0.126 & 0.388 &
             0.780 & 0.193 & 
             0.605 & 0.269 &
             0.454 & 0.245  &
             0.364 & 0.279 &
             0.751 & 0.185 \\  
    LeReS~\cite{leres} & 0.081 & 0.555 &
             0.064 & 0.616 &
             0.278 & 0.427 &
             0.147 & 0.289 &
             \textbf{0.143} & \textbf{0.480} &
             0.145 & 0.503 &
             0.408 & 0.176  &
             0.096 & 0.497 &
             0.241 & 0.325 \\  \hline
    Ours & \textbf{0.042} & \textbf{0.736} &
             \textbf{0.059} & \textbf{0.610} &
             \textbf{0.159} & \textbf{0.485} &
             \textbf{0.050} & \textbf{0.645} &
             0.145 & 0.445 &
             \textbf{0.036} & \textbf{0.814} &
             \textbf{0.069} & \textbf{0.638}  &
             \textbf{0.045} & \textbf{0.700} &
             \textbf{0.060} & \textbf{0.663} \\

    \bottomrule
  \end{tabular}}}
  \label{tab: NYUD reconstruction cmp.}
\end{table*}

\begin{figure*}[]
\centering
\includegraphics[width=0.95\textwidth]{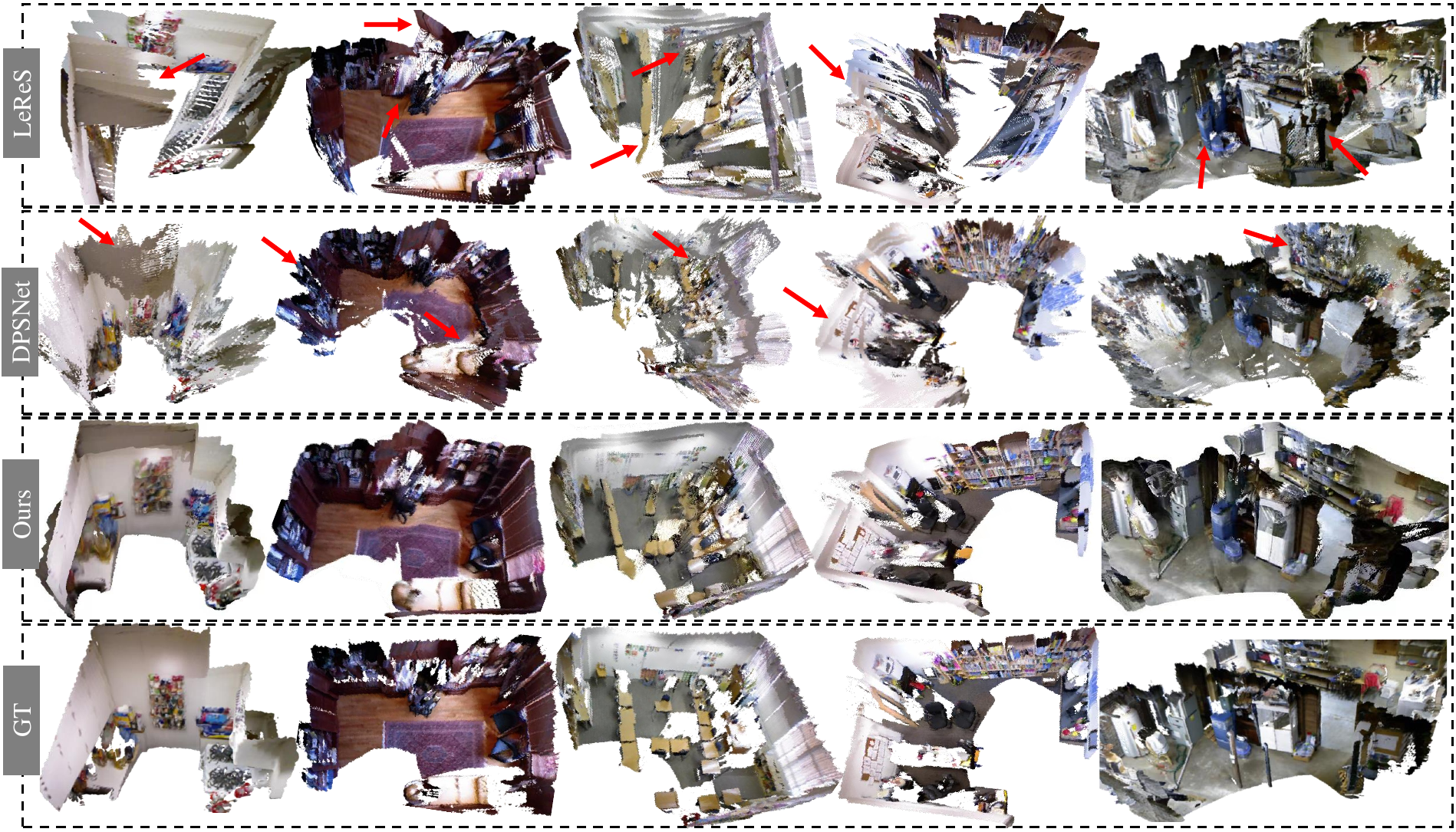}
\vspace{-1 em}
\caption{\textbf{Reconstruction of zero-shot scenes with multiple views.} We sample several NYUv2 scenes for 3D reconstruction comparison. As our method can predict accurate metric depth, thus all frame's predictions are  fused together for scene reconstruction. By contrast, LeReS~\cite{leres}'s depth is up to an unknown scale and shift, which causes noticeable distortions. DPSNet~\cite{im2019dpsnet} is a multi-view stereo method, which cannot work well on low-texture regions. }
\label{fig: visual nyud reconstruction cmp.}
\vspace{-1em}
\end{figure*}

\noindent\textbf{Evaluation on metric depth benchmarks.} To evaluate the accuracy of predicted metric depth, firstly,  we compare with state-of-the-art (SOTA) metric depth prediction methods on NYUv2~\cite{silberman2012indoor}, KITTI~\cite{geiger2012we}.
We use the same model to do %
all evaluations. %
Results are reported in Tab.~\ref{table:errors cmp on NYUD-V2}. Without any fine-tuning or metric adjustment,  we can achieve comparable performance with SOTA methods, which are trained on benchmarks for hundreds of epochs. %

Furthermore, We collect $6$ unseen datasets to do more metric accuracy evaluation. These datasets contain a wide range of indoor and outdoor scenes, including rooms, buildings, and driving scenes. The camera models are also various, e.g. 7scenes has a short focal length (around 500), while ETH3D is 2000. We mainly compare with the SOTA metric depth estimation methods and take their NYUv2 and KITTI models for indoor and outdoor scenes evaluation respectively. From Tab. \ref{table: metric eval on more datasets.}, we observe that although 7Scenes is similar to NYUv2 and NuScenes is similar to KITTI, existing methods face a noticeable performance decrease. In contrast, our model is more robust. %

\begin{table*}[]
\centering
 \caption{Quantitative comparison with SOTA metric depth methods on $6$ unseen benchmarks. For SOTA methods, we use their NYUv2 and KITTI models for indoor and outdoor scenes evaluation respectively, while we use the same model for all zero-shot testing. }
 \vspace{-1 em}
 \resizebox{0.9\linewidth}{!}{%
\begin{tabular}{l|lll|lll}
\toprule[1pt]
\multirow{2}{*}{Method}        & DIODE(Indoor) & iBIMS-1 & 7Scenes      & DIODE(Outdoor)      & ETH3D      & NuScenes     \\
        & \multicolumn{3}{c|}{Indoor scenes (AbsRel$\downarrow$/RMS$\downarrow$)}  & \multicolumn{3}{c}{Outdoor scenes (AbsRel$\downarrow$/RMS$\downarrow$)} \\ \hline
Adabins~\cite{bhat2021adabins}  
         &  0.443 / 1.963       
         &0.212 / 0.901         
         & 0.218 / 0.428 
         &0.865 / 10.35                     
         &1.271 / 6.178            
         &0.445 / 10.658              \\
NewCRFs~\cite{yuan2022new}  
         &0.404 / 1.867               
         &0.206 / 0.861         
         &0.240 / 0.451 
         &0.854 / 9.228                     
         &0.890 / 5.011            
         &0.400 / 12.139              \\ \hline
Ours\_CSTM\_label    
        &\textbf{0.252} / \underline{1.440}               
        & \underline{0.160} / \textbf{0.521}         
        &  \textbf{0.183} / \textbf{0.363}   
        &\textbf{0.414} / \underline{6.934}                     
        & \underline{0.416} / \underline{3.017}          
        & \underline{0.154} / \underline{7.097}             \\
Ours\_CSTM\_image    
        & \underline{0.268} / \textbf{1.429}         
        & \textbf{0.144} / \underline{0.646}        
        & \underline{0.189} / \underline{0.388}  
        & \underline{0.535} / \textbf{6.507}                    
        & \textbf{0.342} / \textbf{2.965}           
        & \textbf{0.147} / \textbf{5.889}    \\ \toprule[1pt]
\end{tabular}}
\label{table: metric eval on more datasets.}
\vspace{-1 em}
\end{table*}

\begin{table*}[t]
\centering
\caption{
Comparison with SOTA affine-invariant depth methods on 5 zero-shot transfer benchmarks.
Our model significantly outperforms previous methods and sets new state-of-the-art. Following the benchmark setting, all methods have manually aligned the scale and shift. 
}
\vspace{-1 em}
\setlength{\tabcolsep}{2pt}
\resizebox{0.99\linewidth}{!}{%
\begin{tabular}{ r |ll|ll|ll|ll|ll|ll|l}
\toprule[1pt]
\multirow{2}{*}{Method} & \multirow{2}{*}{Backbone} & \multirow{2}{*}{\#Params} & \multicolumn{2}{c|}{NYUv2} & \multicolumn{2}{c|}{KITTI} & \multicolumn{2}{c|}{DIODE} & \multicolumn{2}{c|}{ScanNet} & \multicolumn{2}{c|}{ETH3D}  & \multicolumn{1}{c}{Rank} \\
&  &   & AbsRel$\downarrow$     & $\delta_{1}$$\uparrow$     & AbsRel$\downarrow$      & $\delta_{1}$$\uparrow$      & AbsRel$\downarrow$      & $\delta_{1}$$\uparrow$      &AbsRel$\downarrow$      & $\delta_{1}$$\uparrow$       &AbsRel$\downarrow$     & $\delta_{1}$$\uparrow$  & \\ \hline
DiverseDepth~\cite{yin2021virtual}& ResNeXt50~\cite{xie2017aggregated}& 25M  
&$0.117$ &$0.875$ 
&$0.190$ &$0.704$ 
&$0.376$ &$0.631$ 
&$0.108$ &$0.882$ 
&$0.228$ &$0.694$  & $7.7$ \\
MiDaS~\cite{Ranftl2020}& ResNeXt101&  88M %
&$0.111$ &$0.885$ 
&$0.236$ &$0.630$ 
&$0.332$ &$0.715$ 
&$0.111$ &$0.886$  
& $0.184$ &$0.752$ & $7.2$ \\
Leres~\cite{leres} & ResNeXt101&  %
&$0.090$  &${0.916}$  
&${0.149}$ &${0.784}$ 
&${0.271}$ &${0.766}$ 
&${0.095}$ &${0.912}$ 
&${0.171}$ &${0.777}$ & $5.4$ \\
Omnidata~\cite{eftekhar2021omnidata} & ViT-base& %
& 0.074 & 0.945 
& 0.149 & 0.835
& 0.339 & 0.742 
& 0.077 & 0.935 
& 0.166 & 0.778 & $4.9$ \\
HDN~\cite{zhang2022hierarchical} & ViT-Large~\cite{dosovitskiy2020an}&  306M  
&$0.069$  &$0.948$  
&$0.115$ &$0.867$ 
&$0.246$ &$0.780$ 
&$0.080$ &$0.939$ 
&$0.121$ &$0.833$ & $3.7$ \\
DPT-large~\cite{ranftl2021vision} & ViT-Large& %
& 0.098 & 0.903 
& 0.10 & 0.901
& \textbf{0.182} & 0.758 
& 0.078 & 0.938 
& 0.078 & 0.946 & $3.8$ \\
\hline
Ours CSTM\_image & ConvNeXt-large~\cite{liu2022convnet}&  198M 
&$\underline{0.058}$  &$\underline{0.963}$  
&$\textbf{0.053}$ &$\underline{0.965}$ 
&$\underline{0.211}$ &$\textbf{0.825}$ 
&$\textbf{0.074}$ &$\textbf{0.942}$ 
&$\textbf{0.064}$ &$\textbf{0.965}$ & $1.3$ \\ 
Ours CSTM\_label & ConvNeXt-large&   
&$\textbf{0.050}$  &$\textbf{0.966}$  
&$\underline{0.058}$ &$\textbf{0.970}$ 
&$0.224$ &$\underline{0.805}$ 
&$\textbf{0.074}$ &$\underline{0.941}$ 
&$\underline{0.066}$ &$\underline{0.964}$ & $1.8$ \\

 \toprule[1pt]
\end{tabular}}

\label{Table: generalization evaluation.}
\vspace{-1.5em}
\end{table*}

\noindent\textbf{Generalization over diverse scenes.}
Affine-invariant depth benchmarks decouple the scale's effect, which aims to evaluate the model's generalization ability to diverse scenes. Recent impact works, such as MiDaS, LeReS, and DPT, achieved promising performance on them. Following them, we test on 5 datasets and manually align the scale and shift to the ground-truth depth before evaluation. Results are reported in Tab.~\ref{Table: generalization evaluation.}. Although our method enforces the network to recover more challenging metric information, our method outperforms them by a large margin on most datasets.

\begin{figure}[!bth]
\centering
\includegraphics[width=0.5\textwidth]{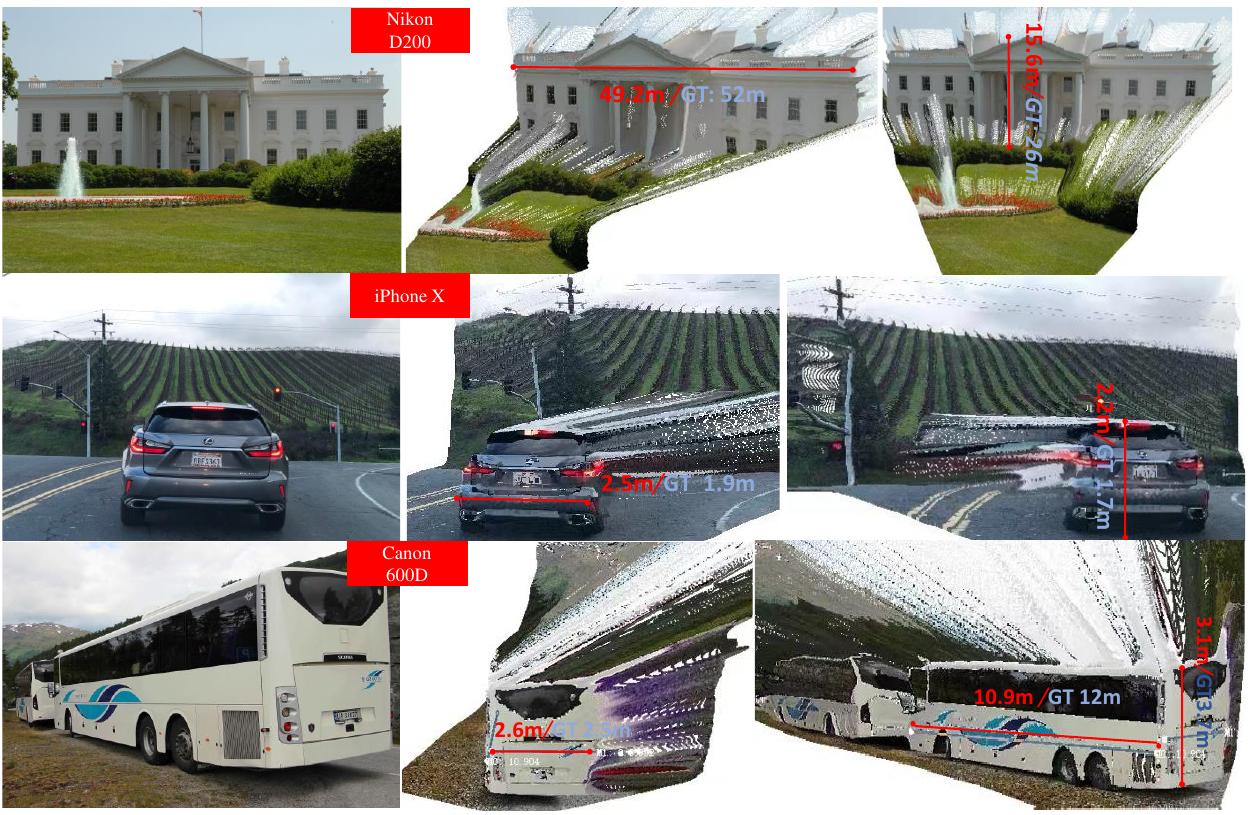}
\vspace{-2 em}
\caption{\textbf{Reconstruction of in-the-wild scenes.} We collect several Flickr photos, which are captured by various cameras. With photos' metadata, we reconstruct the 3D metric shape and measure structures' sizes. Red and blue marks are ours and ground-truth sizes respectively. }
\label{fig: reconstruction in the wild.}
\vspace{-1em}
\end{figure}

\subsection{Applications Based on Our Method}
In these experiments, we apply the CSTM\_image model to various tasks. 

\noindent\textbf{3D scene reconstruction .}
To demonstrate our work can recover the 3D metric shape in the wild, we first do the quantitative comparison on 9 NYUv2 scenes, which are unseen during training. We predict the per-frame metric depth and then fuse them together with provided camera poses. Results are reported in Tab. \ref{tab: NYUD reconstruction cmp.}. We compare with the video consistent depth prediction method (RCVD~\cite{kopf2021rcvd}), the unsupervised video depth estimation method (SC-DepthV2~\cite{bian2021tpami}), the 3D scene shape recovery method (LeReS~\cite{leres}), affine-invariant depth estimation method (DPT~\cite{ranftl2021vision}), and the multi-view stereo reconstruction method (DPSNet~\cite{im2019dpsnet}). Apart from DPSNet and our method, other methods have to align the scale with the ground truth depth for each frame. Although our method does not aim for the video or multi-view reconstruction problem, our method can achieve promising consistency between frames and reconstruct much more accurate 3D scenes than others on these zero-shot scenes.  From the qualitative comparison in Fig.~\ref{fig: visual nyud reconstruction cmp.}. our reconstructions have much less noise and outliers. 

\noindent\textbf{Dense-SLAM mapping.}
Monocular SLAM is an important robotics application. It only relies on a monocular video input to create the trajectory and dense 3D mapping. Owing to limited photometric and geometric constraints, existing methods face serious scale drift problems in large scenes and cannot recover the metric information. Our robust metric depth estimation method is a strong depth prior to the SLAM system. To demonstrate this benefit,  we naively input our metric depth to the SOTA SLAM system, Droid-SLAM~\cite{teed2021droid}, and evaluate the trajectory on KITTI. We do not do any tuning on the original system. Trajectory comparisons are reported in Tab. \ref{tab: KITTI SLAM.}. As Droid-SLAM can access accurate per-frame metric depth, like an RGB-D SLAM, the translation drift ($t_{rel}$) decreases significantly. Furthermore, with our depths, Droid-SLAM can perform denser and more accurate 3D mapping. An example is shown in Fig.~\ref{Fig: first page fig.} and more cases are shown in the supplementary materials.   

We also test on the ETH3D SLAM benchmarks. Results are reported in Tab.~\ref{Tab: ETH3D SLAM}. Droid with our depths has much better SLAM performance. As the ETH3D scenes are all small-scale indoor scenes, the performance improvement is less than that on KITTI. 

\begin{table}[]
\renewcommand\arraystretch{1.1}
\caption{Comparison with SOTA SLAM methods on KITTI. We input predicted metric depth to the Droid-SLAM~\cite{teed2021droid} (`Droid+Ours'), which outperforms others by a large margin on trajectory accuracy.}
\centering
\resizebox{.98\linewidth}{!}{%
  \centering
  \small 
  \setlength{\tabcolsep}{0.5mm}{\begin{tabular}{@{} l |c|c|c|c|c|c|c@{}}
    \toprule
    \multirow{2}{*}{Method} & Seq 00 & Seq 02 & Seq 05 & Seq 06 & Seq 08 & Seq 09 & Seq 10 \\ \cline{2-8} 
      & \multicolumn{7}{c}{Translational RMS drift ($t_{rel}, \downarrow$) / Rotational RMS drift ($r_{rel}, \downarrow$)} \\ \hline
    GeoNet~\cite{yin2018geonet} & 27.6/5.72 & 42.24/6.14 & 
             20.12/7.67 & 
             9.28/4.34 & 
             18.59/7.85 & 
             23.94/9.81 & 
             20.73/9.1  \\
    VISO2-M~\cite{song2015high}  & 12.66/2.73 & 
             9.47/1.19 & 
             15.1/3.65 & 
             6.8/1.93 & 
             14.82/2.52 & 
             3.69/1.25 & 
             21.01/3.26  \\

    ORB-V2~\cite{murORB2} & 11.43/0.58 & 
             10.34/0.26 &
             9.04/0.26 & 
             14.56/0.26 & 
             11.46/0.28 & 
             9.3/0.26 & 
             2.57/0.32    \\
              
    Droid~\cite{teed2021droid} & 33.9/\textbf{0.29} & 
             34.88/\textbf{0.27} & 
             23.4/0.27 & 
             17.2/0.26 & 
             39.6/0.31 & 
             21.7/0.23 & 
             7/0.25   \\  \hline
             
    Droid+Ours & \textbf{1.44}/0.37 & 
             \textbf{2.64}/0.29 & 
             \textbf{1.44}/\textbf{0.25} & 
             \textbf{0.6}/\textbf{0.2} & 
             \textbf{2.2}/\textbf{0.3} & 
             \textbf{1.63}/\textbf{0.22} & 
             \textbf{2.73}/\textbf{0.23}    \\

    \bottomrule
  \end{tabular}}}
  \label{tab: KITTI SLAM.}
\end{table}

\begin{table}[t]
\caption{
Comparison of VO error on ETH3D benchmark. Droid SLAM system is input with our depth (`Droid + Ours'), and ground-truth depth (`Droid + GT'). The average trajectory error is reported.
}
 \resizebox{\linewidth}{!}{%
\begin{tabular}{l|llllll}
\hline
             & Einstein\_global & Manquin4 & Motion1 & Plantscene3 & sfm\_house\_loop & sfm\_lab\_room2 \\ \hline
& \multicolumn{6}{c}{Average trajectory error ($\downarrow$)}  \\ \hline
Droid        & 4.7                               & 0.88     & 0.83    & 0.78        & 5.64             & 0.55            \\ 
Droid + Ours & 1.5                               & 0.69     & 0.62    & 0.34        & 4.03             & 0.53            \\ 
Droid + GT   & 0.7                               & 0.006    & 0.024   & 0.006       & 0.96             & 0.013           \\ \hline
\end{tabular}}
\label{Tab: ETH3D SLAM}
\vspace{-1 em}
\end{table}

\noindent\textbf{Metrology in the wild.} To show the robustness and accuracy of our recovered metric 3D, we download Flickr photos captured by various cameras and collect coarse camera intrinsic parameters from their metadata. We use our CSTM\_image model to reconstruct their metric shape and measure structures' sizes (marked in red in Fig.~\ref{fig: reconstruction in the wild.}), while the ground-truth sizes are in blue. It shows that our measured sizes are very close to the ground-truth sizes.

\subsection{Ablation Study}
\noindent\textbf{Ablation on canonical transformation.}
We study the effect of our proposed canonical transformation for the input images (`CSTM\_input') and the canonical transformation for the ground-truth labels (`CSTM\_output'). Results are reported in  Tab. \ref{table: importance of camera model.}. We train the model on sampled mixed data (55K images) and test it on 6 datasets. A naive baseline (`Ours w/o CSTM') is to remove CSTM modules and enforce the same supervision as ours. Without CSTM, the model is unable to converge when training on mixed metric datasets and cannot achieve metric prediction ability on zero-shot datasets. This is why recent mixed-data training methods compromise learning the affine-invariant depth to avoid metric issues. In contrast, our two CSTM methods both can enable the model to achieve the metric prediction ability, and they can achieve comparable performance. Tab. \ref{table:errors cmp on NYUD-V2} also shows comparable performance. Therefore, both adjusting the supervision and the input image appearance during training can solve the metric ambiguity issues. Furthermore, we compare with CamConvs~\cite{facil2019cam}, which encodes the camera model in the decoder with a 4-channel feature. `CamConvs' employ the same training schedule, model, and training data as ours. This method enforces the network to implicitly understand various camera models from the image appearance and then bridges the imaging size to the real-world size. We believe that this method challenges the data diversity and network capacity, thus their performance is worse than ours.

\begin{table}[]
\caption{Effectiveness of our CSTM. CamConvs~\cite{facil2019cam} directly encodes various camera models in the network, while we perform a simple yet effective transformation to solve the metric ambiguity. Without CSTM, the model cannot achieve transferable metric prediction ability.}
\vspace{-1 em}
\scalebox{0.65}{
\begin{tabular}{l|lll|lll}
\toprule[1pt]
\multirow{2}{*}{Method}        & DDAD & Lyft & DS & NS & KITTI & NYU \\ 
        &\multicolumn{3}{c|}{Test set of train. data (AbsRel$\downarrow$)}     & \multicolumn{3}{c}{Zero-shot test set (AbsRel$\downarrow$)} \\  \hline 
w/o CSTM &$0.530$ &$0.582$  &$0.394$  &$1.00$ & $0.568$      &$0.584$  \\
CamConvs~\cite{facil2019cam}  &$0.295$ &$0.315$  &$0.213$ &$0.423$  &$0.178$      &$0.333$   \\
Ours CSTM\_image &$0.190$ &$0.235$  &$0.182$  &$0.197$ & $0.097$      &$0.210$ \\
Ours CSTM\_label &$0.183$ &$0.221$  &$0.201$  &$0.213$ & $0.081$      &$0.212$  \\
\toprule[1pt]
\end{tabular}}
\label{table: importance of camera model.}
\vspace{-1 em}
\end{table}

\begin{figure}[]
\centering
\includegraphics[width=0.3\textwidth]{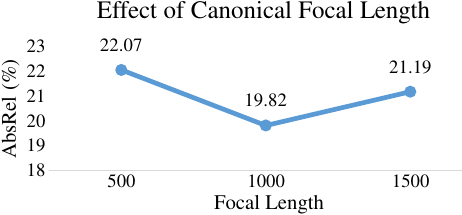}
\caption{\textbf{Effect of different canonical focal lengths.} We experiment on different canonical focal lengths and find that too large or small focal lengths will impact the performance. }
\label{fig: canonical focal length.}
\end{figure}

\noindent\textbf{Ablation on canonical space.}
We study the effect of the canonical camera here, \textit{i.e.}, the canonical focal length. We train the model on the small sampled dataset and test it on the validation set of training data and testing data. The average AbsRel error is calculated.  We experiment on 3 different focal lengths, \ie, 500, 1000, 1500. Experiments show that $focal=1000$ has slightly better performance than others, see Fig.~\ref{fig: canonical focal length.} for details. Thus we set the canonical focal length to 1000 in our experiments.

\noindent\textbf{Effectiveness of 
the random proposal normalization loss.}
To show the effectiveness of our proposed random proposal normalization loss (RPNL), we experiment on the sampled small dataset. Results are shown in Tab.~\ref{table: effectiveness of rpnl.}. We test on the DDAD, Lyft, DrivingStereo (DS), NuScenes (NS), KITTI, and NYUv2.  The `baseline' employs all losses except our RPNL. We compare it with `baseline + RPNL' and `baseline + SSIL~\cite{Ranftl2020}'. We can observe that our proposed random proposal normalization loss can further improve the performance. 
In 
contrast, the scale-shift invariant loss~\cite{Ranftl2020}, which does the normalization on the whole image, can only slightly improve the performance. 
\begin{table}[]
\caption{Effectiveness of random proposal normalization loss. Baseline is supervised by `$L_{\PWN} + L_{\VNL} + L_{silog}$'. SSIL is the scale-shift invariant loss proposed in ~\cite{Ranftl2020}.}
\vspace{-1 em}
\scalebox{0.65}{
\begin{tabular}{l|lll|lll}
\toprule[1pt]
\multirow{2}{*}{Method}        & DDAD & Lyft & DS & NS & KITTI & NYUv2 \\ 
        &\multicolumn{3}{c|}{Test set of train. data (AbsRel$\downarrow$)}     & \multicolumn{3}{c}{Zero-shot test set (AbsRel$\downarrow$)} \\  \hline 
baseline  &$0.204$ &$0.251$  &$0.184$  &$0.207$ &$0.104$      &$0.230$     \\
baseline + SSIL~\cite{Ranftl2020} &$0.197$ &$0.263$  &$0.259$  &$0.206$ & $0.105$      &$0.216$     \\
baseline + RPNL   &$\textbf{0.190}$  &$\textbf{0.235}$  &$\textbf{0.182}$  &$\textbf{0.197}$ &$\textbf{0.097}$      &$\textbf{0.210}$     \\  \toprule[1pt]
\end{tabular}}
\label{table: effectiveness of rpnl.}
\vspace{-2 em}
\end{table}

\section{Conclusion} In this paper, we 
tackle 
the problem of reconstructing the 3D metric scene from a single monocular image. To solve the depth ambiguity in image appearance caused by various focal lengths, we propose a canonical camera space transformation method. With our method, we can easily merge millions of data captured by 10k cameras to train one metric depth model. To improve the robustness, we collected over $8$M data for training. Several zero-shot evaluations show the effectiveness and robustness of our work. We further show the ability to do metrology on randomly collected internet images and dense mapping on large-scale scenes. 

\section*{Acknowledgements}

This work was in part supported by National Key R\&D Program of China (No.\  2022ZD0118700).

\section{Appendix}

\subsection{Datasets and Training and Testing}
We collect over $8$M data from 11 public datasets for training. Datasets are listed in Tab.~\ref{table: datasets}. The autonomous driving datasets, including DDAD~\cite{packnet}, Lyft~\cite{lyftl5preception}, DrivingStereo~\cite{yang2019drivingstereo}, Argoverse2~\cite{Argoverse2}, DSEC~\cite{Gehrig21ral}, and Pandaset~\cite{itsc21pandaset}, have provided LiDar and camera intrinsic and extrinsic parameters. We project the LiDar to image planes to obtain ground-truth depths. In contrast, Cityscapes~\cite{Cordts2016Cityscapes}, DIML~\cite{cho2021diml}, and UASOL~\cite{bauer2019uasol} only provide calibrated stereo images. We use draftstereo~\cite{lipson2021raft} to achieve pseudo ground-truth depths. Mapillary PSD~\cite{MapillaryPSD} dataset provides paired RGB-D, but the depth maps are achieved from a structure-from-motion method. The camera intrinsic parameters are estimated from the SfM. We believe that such achieved metric information is noisy. Thus we do not enforce learning-metric-depth loss on this data, \textit{i.e.},  $L_{silog}$, to reduce the effect of noises. For the Taskonomy~\cite{zamir2018taskonomy} dataset, we follow LeReS~\cite{yin2022towards} to obtain the instance planes, which are employed in the pair-wise normal regression loss. During training, we employ the training strategy from \cite{yin2020diversedepth_old} to balance all datasets in each training batch. 

The testing data is listed in Tab.~\ref{table: datasets}. All of them are captured by high-quality sensors. In testing, we employ their provided camera intrinsic parameters to perform our proposed canonical space transformation.

\begin{table}[]
\caption{Training and testing datasets used in experiments.}
\vspace{-1 em}
\centering
\begin{threeparttable}
\scalebox{0.8}{
\begin{tabular}{ r llll}
 \toprule[1pt]
\multicolumn{1}{l|}{Datasets}                       & \multicolumn{1}{l|}{Scenes}         & \multicolumn{1}{l|}{%
Label}              & \multicolumn{1}{l|}{Size}        & \# Cam.  \\ \hline
\multicolumn{5}{c}{Training Data}                                                                                                                                                          \\ \hline
\multicolumn{1}{l|}{DDAD~\cite{packnet}}                           & \multicolumn{1}{l|}{Outdoor}        & \multicolumn{1}{l|}{LiDar}                 & \multicolumn{1}{l|}{$\sim$80K}   & 36+            \\
\multicolumn{1}{l|}{Lyft~\cite{lyftl5preception}}                           & \multicolumn{1}{l|}{Outdoor}        & \multicolumn{1}{l|}{LiDar}                 & \multicolumn{1}{l|}{$\sim$50K}   & 6+             \\
\multicolumn{1}{l|}{Driving Stereo (DS)~\cite{yang2019drivingstereo}}                 & \multicolumn{1}{l|}{Outdoor}        & \multicolumn{1}{l|}{Stereo\tnote{\dag}}       & \multicolumn{1}{l|}{$\sim$181K}  & 1              \\
\multicolumn{1}{l|}{DIML~\cite{cho2021diml}}                           & \multicolumn{1}{l|}{Outdoor}        & \multicolumn{1}{l|}{Stereo\tnote{\dag}}       & \multicolumn{1}{l|}{$\sim$122K}  & 10             \\
\multicolumn{1}{l|}{Arogoverse2~\cite{Argoverse2}}                    & \multicolumn{1}{l|}{Outdoor}        & \multicolumn{1}{l|}{LiDar}                 & \multicolumn{1}{l|}{$\sim$3515K} & 6+             \\
\multicolumn{1}{l|}{Cityscapes~\cite{Cordts2016Cityscapes}}                     & \multicolumn{1}{l|}{Outdoor}        & \multicolumn{1}{l|}{Stereo\tnote{\dag}}       & \multicolumn{1}{l|}{$\sim$170K}  & 1              \\
\multicolumn{1}{l|}{DSEC~\cite{Gehrig21ral}}                           & \multicolumn{1}{l|}{Outdoor}        & \multicolumn{1}{l|}{LiDar}                 & \multicolumn{1}{l|}{$\sim$26K}   & 1              \\
\multicolumn{1}{l|}{Mapillary PSD~\cite{MapillaryPSD}} & \multicolumn{1}{l|}{Outdoor}        & \multicolumn{1}{l|}{SfM\tnote{\ddag}} & \multicolumn{1}{l|}{750K}        & 1000+          \\
\multicolumn{1}{l|}{Pandaset~\cite{itsc21pandaset}}                       & \multicolumn{1}{l|}{Outdoor}        & \multicolumn{1}{l|}{LiDar}                 & \multicolumn{1}{l|}{$\sim$48K}         & 6              \\
\multicolumn{1}{l|}{UASOL~\cite{bauer2019uasol}}                          & \multicolumn{1}{l|}{Outdoor}        & \multicolumn{1}{l|}{Stereo\tnote{\dag}}       & \multicolumn{1}{l|}{$\sim$137K}        & 1              \\
\multicolumn{1}{l|}{Taskonomy~\cite{zamir2018taskonomy}}                      & \multicolumn{1}{l|}{Indoor}         & \multicolumn{1}{l|}{LiDar}                 & \multicolumn{1}{l|}{$\sim$4M}       & $\sim$1M            \\ \hline
\multicolumn{5}{c}{Testing Data}                                                                                                                                                           \\ \hline
\multicolumn{1}{l|}{NYU~\cite{silberman2012indoor}}                            & \multicolumn{1}{l|}{Indoor}         & \multicolumn{1}{l|}{Kinect}                & \multicolumn{1}{l|}{654}         & 1              \\
\multicolumn{1}{l|}{KITTI~\cite{Geiger2013IJRR}}                          & \multicolumn{1}{l|}{Outdoor}        & \multicolumn{1}{l|}{LiDar}                 & \multicolumn{1}{l|}{652}           & 4         \\
\multicolumn{1}{l|}{ScanNet~\cite{dai2017scannet}}                        & \multicolumn{1}{l|}{Indoor}         & \multicolumn{1}{l|}{Kinect}                & \multicolumn{1}{l|}{700}         & 1              \\
\multicolumn{1}{l|}{NuScenes (NS)~\cite{caesar2020nuscenes}}                       & \multicolumn{1}{l|}{Outdoor}        & \multicolumn{1}{l|}{LiDar}                 & \multicolumn{1}{l|}{10K}           & 6              \\
\multicolumn{1}{l|}{ETH3D~\cite{schops2017multi}}                          & \multicolumn{1}{l|}{Outdoor}        & \multicolumn{1}{l|}{LiDar}                 & \multicolumn{1}{l|}{431}           & 1              \\
\multicolumn{1}{l|}{DIODE~\cite{vasiljevic2019diode}}                          & \multicolumn{1}{l|}{In/Out} & \multicolumn{1}{l|}{LiDar}                 & \multicolumn{1}{l|}{771}           & 1           \\
\multicolumn{1}{l|}{7Scenes~\cite{shotton2013scene}}                        & \multicolumn{1}{l|}{Indoor}         & \multicolumn{1}{l|}{Kinect}                & \multicolumn{1}{l|}{17k}           & 1              \\
\multicolumn{1}{l|}{iBims-1~\cite{koch2018evaluation}}                        & \multicolumn{1}{l|}{Indoor}         & \multicolumn{1}{l|}{LiDar}                & \multicolumn{1}{l|}{100}           & 1              \\
\toprule[1pt]
\end{tabular}}
\begin{tablenotes}
\footnotesize
\item[\dag]`Stereo': we use RaftStereo~\cite{lipson2021raft} to retrieve the pseudo ground truth.
\item[\ddag]`SfM': pseudo ground truth is retrieved by structure from motion.
\end{tablenotes}
\end{threeparttable}
\label{table: datasets}
\end{table}

\subsection{Details for Some Experiments}
\noindent\textbf{Evaluation of zero-shot 3D scene reconstruction.} In this experiment, we use all methods' released models to predict each frame's depth and use the ground-truth poses and camera intrinsic parameters to reconstruct point clouds. When evaluating the reconstructed point cloud, we employ the iterative closest point (ICP)~\cite{besl1992method} algorithm to match the predicted point clouds with ground truth by a pose transformation matrix. Finally, we evaluate the Chamfer $\ell_1$ distance and F-score on the point cloud.

\begin{figure*}[]
\centering
\includegraphics[width=0.8\textwidth]{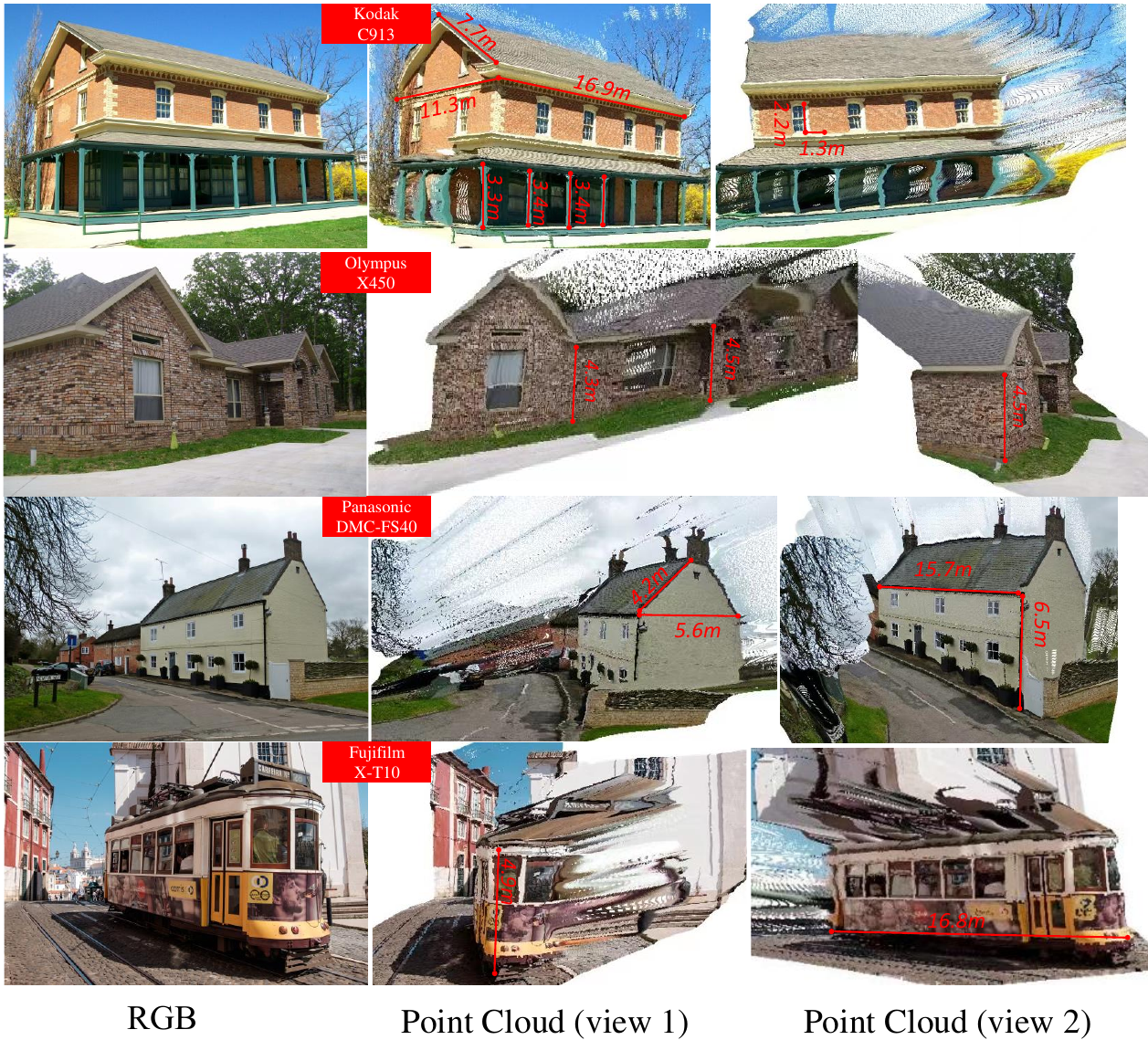}
\caption{\textbf{3D metric reconstruction of in-the-wild images.} We collect several Flickr images and use our model to reconstruct the scene. The focal length information is collected from the photo's metadata. From the reconstructed point cloud, we can measure some structures' sizes. We can observe that sizes are in a reasonable range. }
\label{fig: recon in the wild.}
\end{figure*}

\noindent\textbf{Reconstruction of in-the-wild scenes.} We collect several photos from Flickr. From their associated camera metadata, we can obtain the focal length $\hat{f}$ and the pixel size $\delta$. According to $\nicefrac{\hat{f}}{\delta}$, we can obtain the pixel-represented focal length for 3D reconstruction and achieve the metric information. We use meshlab software to measure some structures' size on point clouds. More visual results are shown in Fig. \ref{fig: recon in the wild.}.

\noindent\textbf{Generalization of metric depth estimation.} To evaluate our method's robustness of metric recovery, we test on 8 zero-shot datasets, i.e. NYU, KITTI, DIODE (indoor and outdoor parts), ETH3D, iBims-1, NuScenes, and 7Scenes. Details are reported in Tab.~\ref{table: datasets}.  We use the officially provided focal length to predict the metric depths. All benchmarks use the same depth model for evaluation. We don't perform any scale alignment.  

\noindent\textbf{Evaluation on affine-invariant depth benchmarks.} We follow existing affine-invariant depth estimation methods to evaluate 5 zero-shot datasets. Before evaluation, we employ the least square fitting to align the scale and shift with ground truth~\cite{leres}. Previous methods' performance is cited from their papers. 

\begin{figure*}[]
\centering
\includegraphics[width=0.95\textwidth]{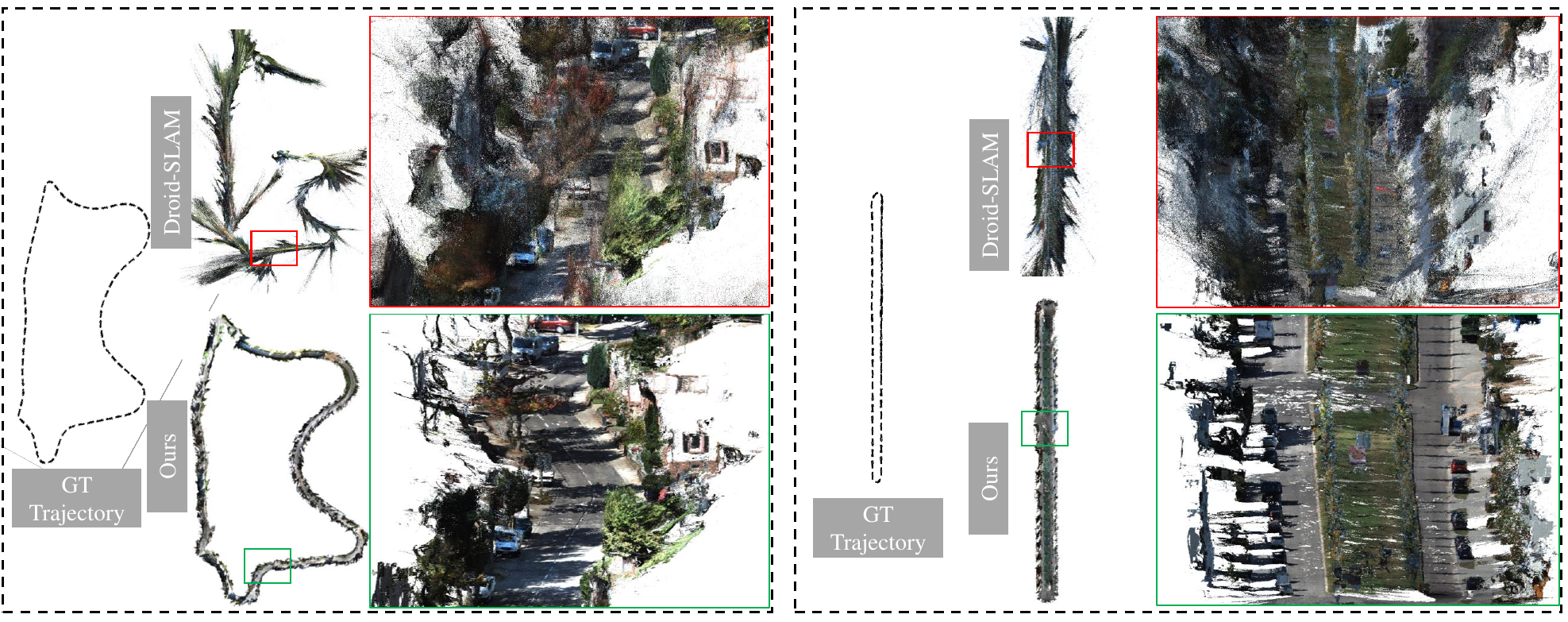}
\includegraphics[width=0.95\textwidth]{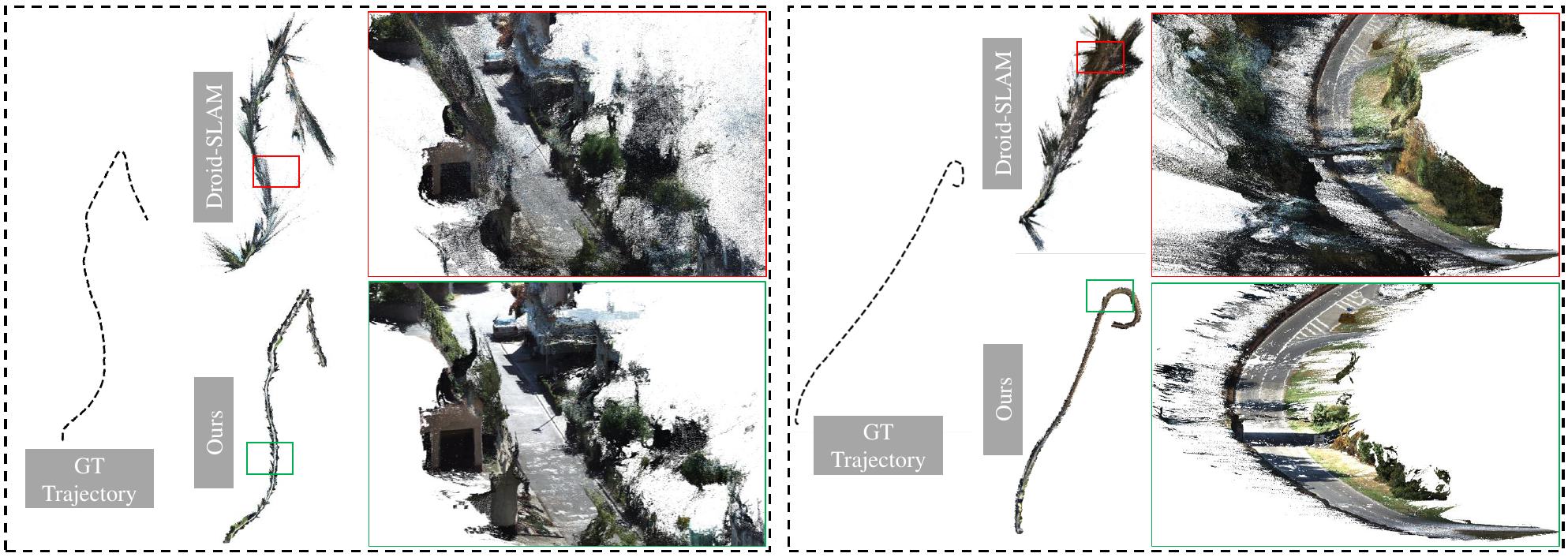}
\caption{\textbf{Dense-SLAM Mapping.} Existing SOTA mono-SLAM methods usually face scale drift problems in large-scale scenes and are unable to achieve the metric scale. We show the ground-truth trajectory and Droid-SLAM~\cite{teed2021droid} predicted trajectory and their dense mapping. Then, we naively input our metric depth to Droid-SLAM, which can recover a much more accurate trajectory and perform the \textit{metric} dense mapping. }
\label{fig: dense_slam1.}
\end{figure*}

\noindent\textbf{Dense-SLAM Mapping.} This experiment is conducted on the KITTI odometry benchmark. We use our model to predict metric depths, and then naively input them to the Droid-SLAM system as an initial depth. We do not perform any finetuning but directly run their released codes on KITTI. With Droid-SLAM predicted poses, we unproject depths to the 3D point clouds and fuse them together to achieve dense metric mapping. More qualitative results are shown in Fig.~\ref{fig: dense_slam1.}.

\subsection{More Visual Results}
\begin{figure*}[]
\centering
\includegraphics[width=1\textwidth]{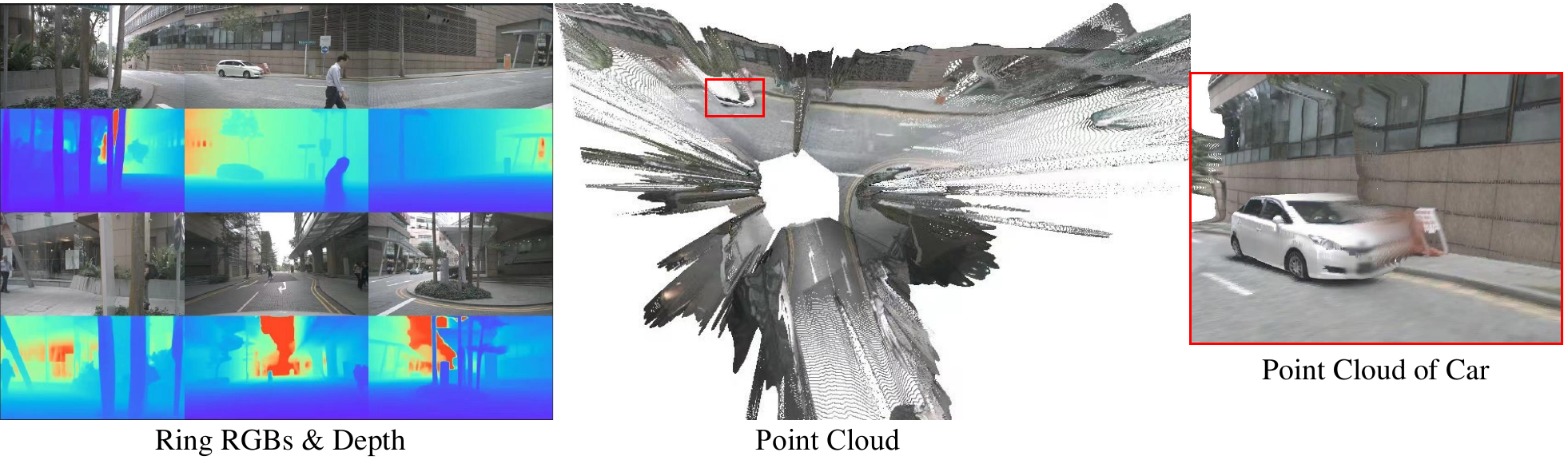}
\caption{\textbf{3D reconstruction of 360\degree views.} Current autonomous driving cars are equipped with several pin-hole cameras to capture 360\degree views. With our model, we can reconstruct each view and smoothly fuse them together. We can see that all views can be well merged together without scale inconsistency problems. Testing data are from NuScenes. Note that the front view camera has a different focal length from other views. }
\label{fig: recon nuscenes.}
\end{figure*}

\noindent\textbf{Reconstructing 360\degree NuScenes scenes.} Current autonomous driving cars are equipped with several pin-hole cameras to capture 360\degree views. Capturing the surround-view depth is important for autonomous driving.  We sampled some scenes from the testing data of NuScenes. With our depth model, we can obtain the metric depths for 6-ring cameras. With the provided camera intrinsic and extrinsic parameters, we unproject the depths to the 3D point cloud and merge all views together. See Fig. \ref{fig: recon nuscenes.} for details. Note that 6-ring cameras have different camera intrinsic parameters. We can observe that all views' point clouds can be fused together consistently.

\begin{figure*}[]
\centering
\includegraphics[width=1\textwidth]{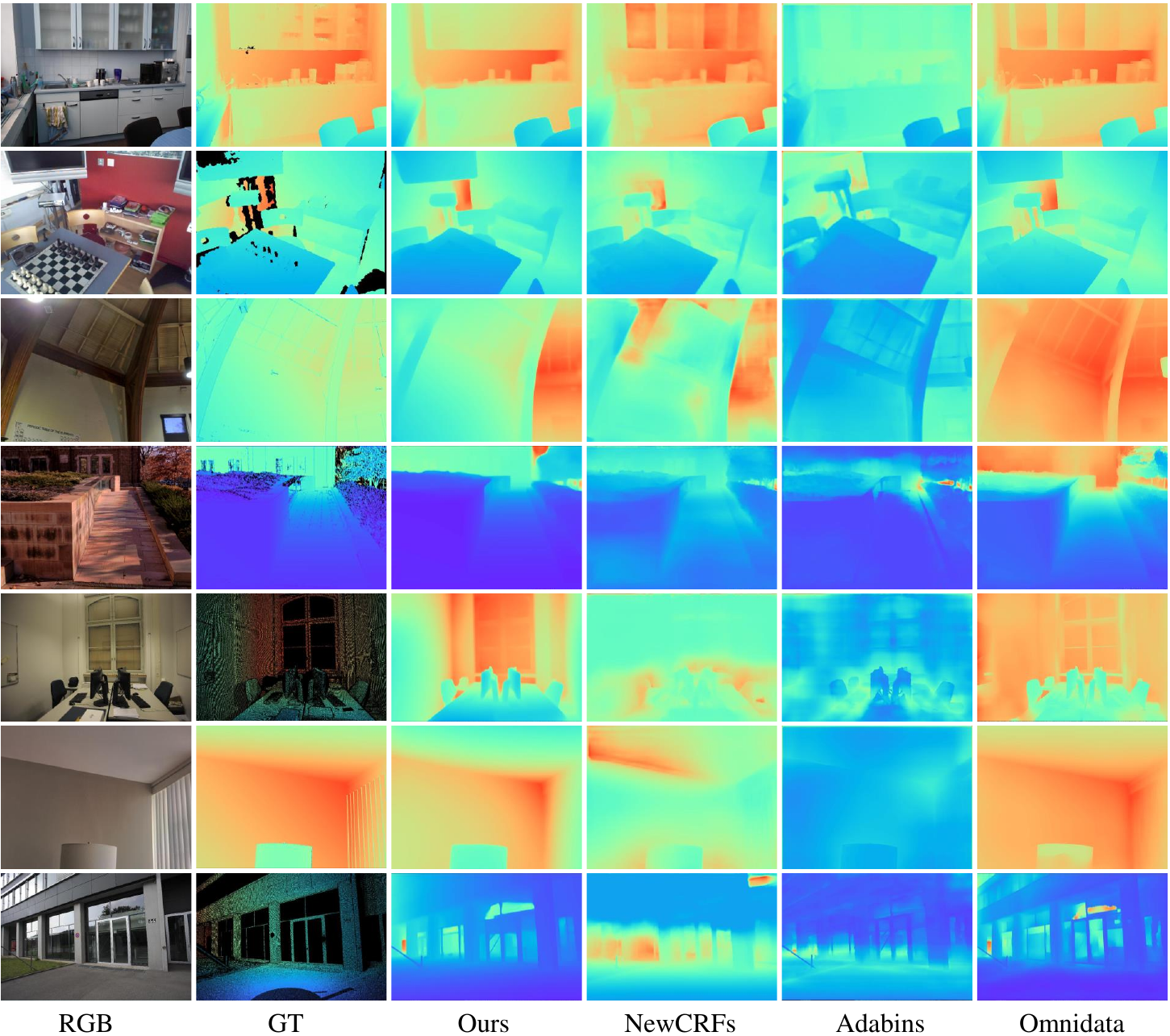}
\caption{\textbf{Depth estimation.} The visual comparison of predicted on iBims, ETH3D, and DIODE.}
\label{fig: depth_cmp1.}
\end{figure*}

\begin{figure*}[]
\centering
\includegraphics[width=0.9\textwidth]{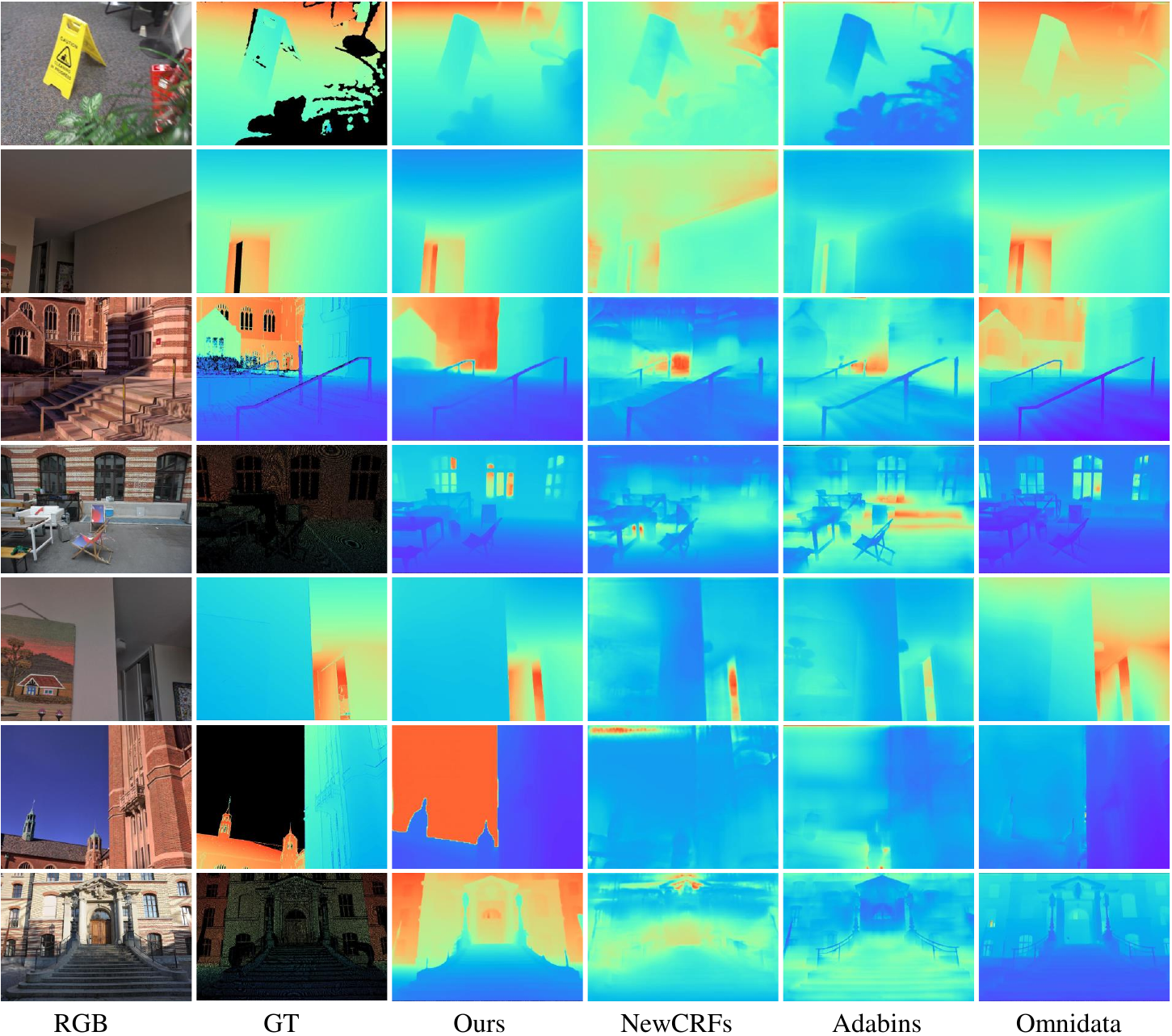}
\caption{\textbf{Depth estimation.} The visual comparison of predicted on iBims, ETH3D, and DIODE.}
\label{fig: depth_cmp2.}
\end{figure*}

\begin{figure*}[]
\centering
\includegraphics[width=0.9\textwidth]{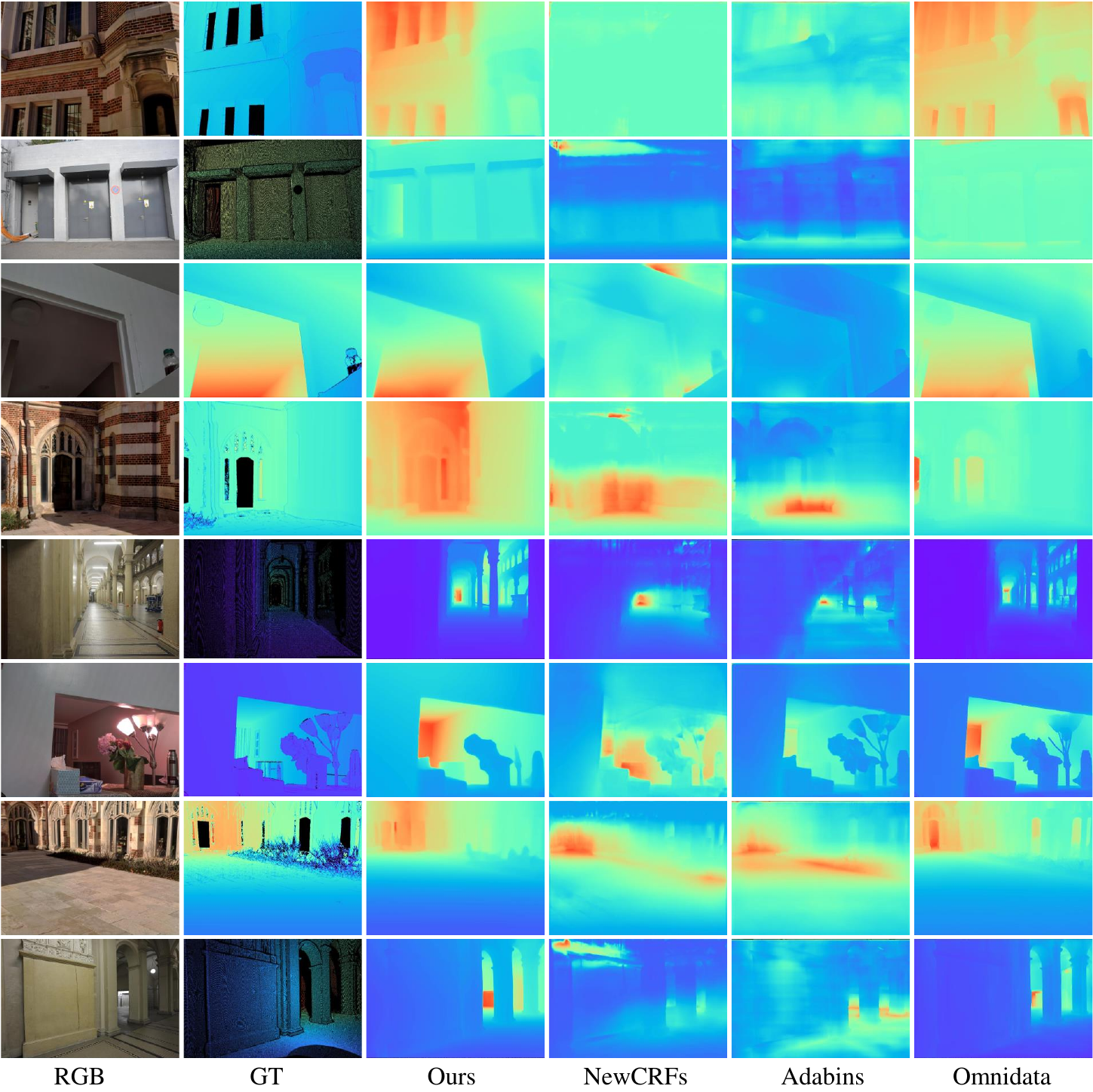}
\caption{\textbf{Depth estimation.} The visual comparison of predicted on iBims, ETH3D, and DIODE.}
\label{fig: depth_cmp3.}
\end{figure*}

\begin{figure*}[]
\centering
\includegraphics[width=0.9\textwidth]{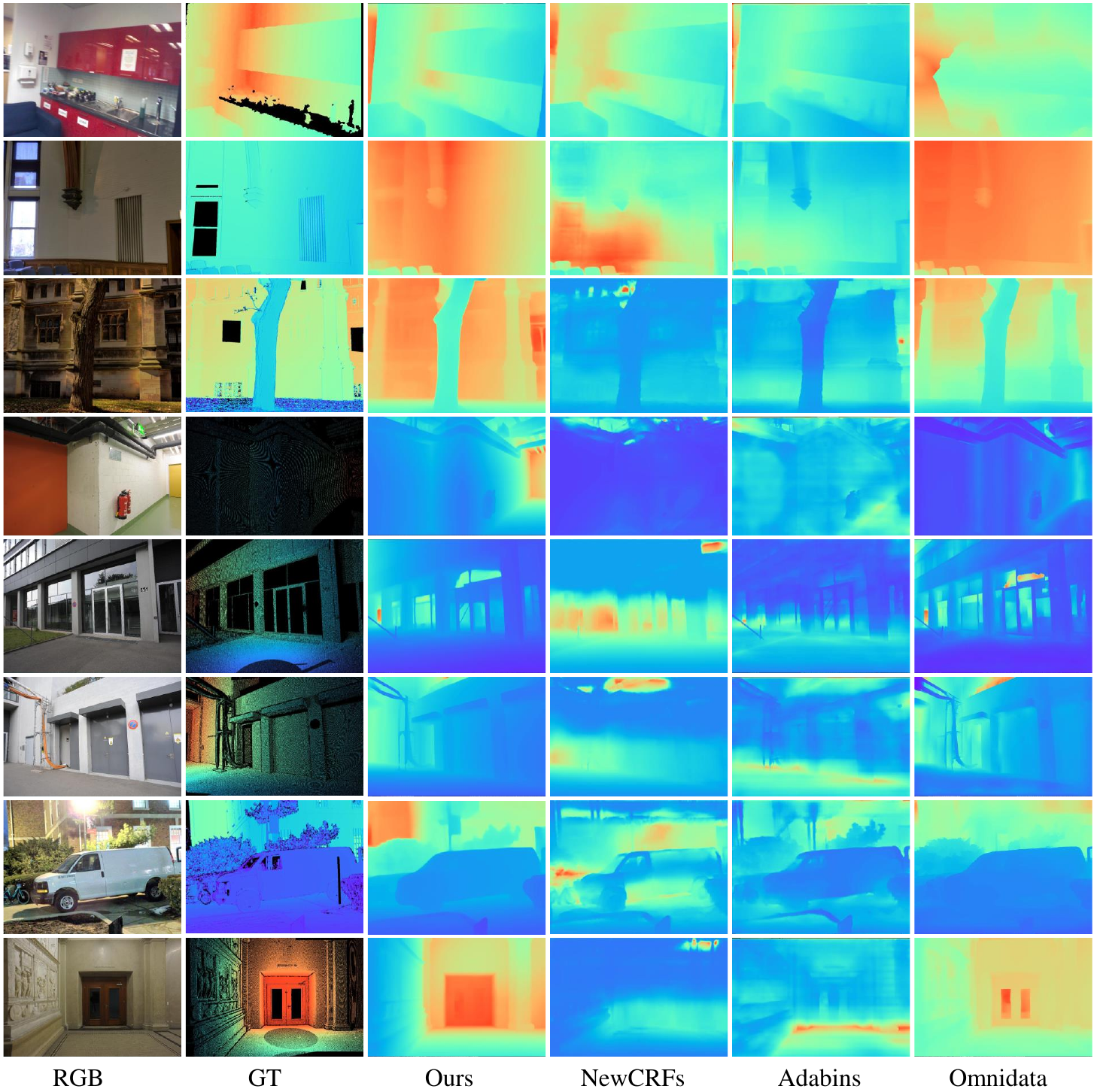}
\caption{\textbf{Depth estimation.} The visual comparison of predicted on iBims, ETH3D, and DIODE.}
\label{fig: depth_cmp4.}
\end{figure*}

\noindent\textbf{Qualitative comparison of depth estimation.} In Figs.~\ref{fig: depth_cmp1.},~\ref{fig: depth_cmp2.},~\ref{fig: depth_cmp3.}, and ~\ref{fig: depth_cmp4.}, We show the qualitative comparison of depth maps with Adabins~\cite{bhat2021adabins}, NewCRFs~\cite{yuan2022new}, and Omnidata~\cite{eftekhar2021omnidata}. Our results have much less artifacts.

{\small
\bibliographystyle{./ICCV2023/ieee_fullname}
\bibliography{./ICCV2023/egbib}
}

\end{document}


\title{Supplementary Materials: Towards Transferable Learning of Single Image Metric 3D in the Wild
}

\author{First Author\\
Institution1\\
Institution1 address\\
{\tt\small firstauthor@i1.org}
\and
Second Author\\
Institution2\\
First line of institution2 address\\
{\tt\small secondauthor@i2.org}
}

\maketitle

\section{Details for Models}
In our work, our encoder employs the convnext-base network, whose pretrained weight is from the official released ImageNet-22k pretraining. The decoder follows the adabins~\cite{bhat2021adabins}. We set the depth bins number to 200, and the depth range is $[0.3m, 120m]$. We establish 4 flip connections from different levels of encoder blocks to the decoder to merge more low-level features. For each flip connection, we concatenate the camera model with the encoder feature. Here, we assume the camera intrinsic parameters are $[f_x, f_y, u0, v0]$. Normally, the calibrated $f_x = f_y = f$. The image size is $H \times W$. We encode the camera model as follows. 
\begin{equation}
\mathbf{u}=\begin{bmatrix}
\frac{0-u_0}{W} &... &\frac{u-u_0}{W} & ... &\frac{W-u_0}{W} \\ 
... &...  &...&...  &... \\ 
\frac{0-u_0}{W} &...  &\frac{v-v_0}{W} & ... &\frac{W-u_0}{W}  \\ 
\end{bmatrix} \in \mathbf{R}^{H \times W}
\end{equation}
\begin{equation}
    \mathbf{v}=\begin{bmatrix}
\frac{0-v_0}{H} &...  &\frac{0-v_0}{H} \\ 
... &...  &... \\ 
\frac{v-v_0}{H} &...  &\frac{v-v_0}{H} \\ 
... &...  &... \\ 
\frac{H-v_0}{H} &...  &\frac{H-v_0}{H} 
\end{bmatrix} \in \mathbf{R}^{H \times W}
\end{equation}
\begin{equation}
    \mathbf{f}_y =  \begin{bmatrix}
arctan(\frac{0-v_0}{f}) &...  &arctan(\frac{0-v_0}{f}) \\ 
... &...  &... \\ 
arctan(\frac{v-v_0}{f}) &...  &arctan(\frac{v-v_0}{f}) \\ 
... &...  &... \\ 
arctan(\frac{H-v_0}{f}) &...  &arctan(\frac{H-v_0}{f})
\end{bmatrix} \in \mathbf{R}^{H \times W}
\end{equation}
\begin{equation}
\mathbf{f}_{x}=\begin{bmatrix}
arctan(\frac{0-u_0}{f}) &...  &arctan(\frac{W-u_0}{f}) \\ 
... &...  &... \\ 
arctan(\frac{0-u_0}{f}) &...    &arctan(\frac{W-u_0}{f}) 
\end{bmatrix} \in \mathbf{R}^{H \times W}
\end{equation}

\section{Details for Datasets and Training}
We collect over $8M$ data from 11 public datasets for training. Datasets list is shown in ~\ref{table: datasets}. The autonomous driving datasets, including DDAD, Lyft, DrivingStereo, Argoverse2, DSEC, and Pandaset, have provided LiDar and camera intrinsic and extrinsic parameters. We project the LiDar to camera image planes to obtain ground-truth depths. In contrast, Cityscapes, DIML, and UASOL do not have ground truth depth, but are with calibrated stereo images. We use draftstereo~\cite{lipson2021raft} to achieve pseudo ground truth. Mapillary PSD dataset provides paired RGB-D, but the depth maps are achieved from a structure-from-motion method. The camera intrinsic parameters are estimated from the SfM. We believe that such achieved metric information is very noisy, so we do not enforce learning-metric-depth loss on this data, i.e. $L_{silog}$, to reduce the effect from noises. For the Taskonomy dataset, we follow LeReS~\cite{yin2022towards} to obtain the instance planes, which is employed in the pair-wise normal regression loss. During training, we employ the training strategy from ~\cite{yin2020diversedepth_old} to balance all datasets in each training batch.

\begin{table}[]
\caption{Training and testing datasets used in experiments.}
\vspace{-1 em}
\centering
\begin{threeparttable}
\scalebox{0.8}{
\begin{tabular}{ r llll}
 \toprule[1pt]
\multicolumn{1}{l|}{Datasets}                       & \multicolumn{1}{l|}{Scenes}         & \multicolumn{1}{l|}{%
Label}              & \multicolumn{1}{l|}{Size}        & \# Cam.  \\ \hline
\multicolumn{5}{c}{Training Data}                                                                                                                                                          \\ \hline
\multicolumn{1}{l|}{DDAD~\cite{packnet}}                           & \multicolumn{1}{l|}{Outdoor}        & \multicolumn{1}{l|}{LiDar}                 & \multicolumn{1}{l|}{$\sim$80K}   & 36+            \\
\multicolumn{1}{l|}{Lyft~\cite{lyftl5preception}}                           & \multicolumn{1}{l|}{Outdoor}        & \multicolumn{1}{l|}{LiDar}                 & \multicolumn{1}{l|}{$\sim$50K}   & 6+             \\
\multicolumn{1}{l|}{Driving Stereo (DS)~\cite{yang2019drivingstereo}}                 & \multicolumn{1}{l|}{Outdoor}        & \multicolumn{1}{l|}{Stereo\tnote{\dag}}       & \multicolumn{1}{l|}{$\sim$181K}  & 1              \\
\multicolumn{1}{l|}{DIML~\cite{cho2021diml}}                           & \multicolumn{1}{l|}{Outdoor}        & \multicolumn{1}{l|}{Stereo\tnote{\dag}}       & \multicolumn{1}{l|}{$\sim$122K}  & 10             \\
\multicolumn{1}{l|}{Arogoverse2~\cite{Argoverse2}}                    & \multicolumn{1}{l|}{Outdoor}        & \multicolumn{1}{l|}{LiDar}                 & \multicolumn{1}{l|}{$\sim$3515K} & 6+             \\
\multicolumn{1}{l|}{Cityscapes~\cite{Cordts2016Cityscapes}}                     & \multicolumn{1}{l|}{Outdoor}        & \multicolumn{1}{l|}{Stereo\tnote{\dag}}       & \multicolumn{1}{l|}{$\sim$170K}  & 1              \\
\multicolumn{1}{l|}{DSEC~\cite{Gehrig21ral}}                           & \multicolumn{1}{l|}{Outdoor}        & \multicolumn{1}{l|}{LiDar}                 & \multicolumn{1}{l|}{$\sim$26K}   & 1              \\
\multicolumn{1}{l|}{Mapillary PSD~\cite{MapillaryPSD}} & \multicolumn{1}{l|}{Outdoor}        & \multicolumn{1}{l|}{SfM\tnote{\ddag}} & \multicolumn{1}{l|}{750K}        & 1000+          \\
\multicolumn{1}{l|}{Pandaset~\cite{itsc21pandaset}}                       & \multicolumn{1}{l|}{Outdoor}        & \multicolumn{1}{l|}{LiDar}                 & \multicolumn{1}{l|}{$\sim$48K}         & 6              \\
\multicolumn{1}{l|}{UASOL~\cite{bauer2019uasol}}                          & \multicolumn{1}{l|}{Outdoor}        & \multicolumn{1}{l|}{Stereo\tnote{\dag}}       & \multicolumn{1}{l|}{$\sim$137K}        & 1              \\
\multicolumn{1}{l|}{Taskonomy~\cite{zamir2018taskonomy}}                      & \multicolumn{1}{l|}{Indoor}         & \multicolumn{1}{l|}{LiDar}                 & \multicolumn{1}{l|}{$\sim$4M}       & $\sim$1M            \\ \hline
\multicolumn{5}{c}{Testing Data}                                                                                                                                                           \\ \hline
\multicolumn{1}{l|}{NYU~\cite{silberman2012indoor}}                            & \multicolumn{1}{l|}{Indoor}         & \multicolumn{1}{l|}{Kinect}                & \multicolumn{1}{l|}{654}         & 1              \\
\multicolumn{1}{l|}{KITTI~\cite{Geiger2013IJRR}}                          & \multicolumn{1}{l|}{Outdoor}        & \multicolumn{1}{l|}{LiDar}                 & \multicolumn{1}{l|}{652}           & 4         \\
\multicolumn{1}{l|}{ScanNet~\cite{dai2017scannet}}                        & \multicolumn{1}{l|}{Indoor}         & \multicolumn{1}{l|}{Kinect}                & \multicolumn{1}{l|}{700}         & 1              \\
\multicolumn{1}{l|}{NuScenes (NS)~\cite{caesar2020nuscenes}}                       & \multicolumn{1}{l|}{Outdoor}        & \multicolumn{1}{l|}{LiDar}                 & \multicolumn{1}{l|}{10K}           & 6              \\
\multicolumn{1}{l|}{ETH3D~\cite{schops2017multi}}                          & \multicolumn{1}{l|}{Outdoor}        & \multicolumn{1}{l|}{LiDar}                 & \multicolumn{1}{l|}{431}           & -\tnote{$\ast$}              \\
\multicolumn{1}{l|}{DIODE~\cite{vasiljevic2019diode}}                          & \multicolumn{1}{l|}{In/Out} & \multicolumn{1}{l|}{LiDar}                 & \multicolumn{1}{l|}{771}           & -\tnote{$\ast$}           \\
\multicolumn{1}{l|}{7Scenes~\cite{shotton2013scene}}                        & \multicolumn{1}{l|}{Indoor}         & \multicolumn{1}{l|}{Kinect}                & \multicolumn{1}{l|}{17k}           & 1              \\\toprule[1pt]
\end{tabular}}
\begin{tablenotes}
\footnotesize
\item[\dag]`Stereo': we use RaftStereo~\cite{lipson2021raft} to retrieve the pseudo ground truth.
\item[\ddag]`SfM': pseudo ground truth is retrieved by structure from motion.
\item[$\ast$] `-': camera intrinsic parameters are unavailable.
\end{tablenotes}
\end{threeparttable}
\label{table: datasets}
\end{table}

\section{Details for Some Experiments}
\noindent\textbf{Evaluation of zero-shot 3D scene reconstruction.} In this experiment, we use all methods' released models to  predict each frame's depth and use the ground truth pose to merge all frames together. When evaluating the reconstructed point cloud, we employ the iterative closest point (ICP)~\cite{besl1992method} algorithm to match the predicted point clouds with ground truth by a pose transformation matrix. Finally, we evaluate the Chamfer l1 distance and F-score on the point cloud.

\noindent\textbf{Reconstruction of in-the-wild scenes.} We collect several photos from Flickr. From their associated camera metadata, we can obtain the focal length $\hat{f}$ and the pixel size $\delta$. With $\nicefrac{\hat{f}}{\delta}$, we can obtain the pixel-represented focal length for reconstruction and achieve the metric information. We use meshlab to measure some structures' size on point clouds. 

\noindent\textbf{Generalization of metric depth estimation.} In these comparisons, we use the official provided focal length to predict the metric depths. All benchmarks use the same depth model for evaluation. We don't perform any scale alignment and do the zero-shot testing on them. 

\noindent\textbf{Generalization over diverse scenes.} We follow existing affine-invariant depth estimation  methods to evaluate on 5 zero-shot datasets. Before evaluation, we employ a least square fitting to align the scale and shift with ground truth~\cite{leres}. Previous methods' performance are cited from their papers.

\section{More Visual Results}
We show more visual results in the supplementary materials. We sampled some scenes from the testing data of DDAD. With our depth model, we can obtain the metric depths for 6 ring cameras. With the provided camera intrinsic and extrinsic parameters, we unproject the depths to the 3D point cloud and merge all views together. See enclosed videos for details. Note that 6 ring cameras have different camera intrinsic parameters.

\maketitle

\def\PWN{{\rm PWN}}
\def\VNL{{\rm VNL}}
\def\RPNL{{\rm RPNL}}

{\small
\bibliographystyle{ieee_fullname}
\bibliography{draft}
}